\documentclass[preprint,12pt,authoryear]{elsarticle}

\usepackage{hyperref}
\usepackage{amssymb}
\usepackage{amsmath}
\usepackage{booktabs}
\usepackage{multirow}
\usepackage{makecell}
\usepackage{threeparttable}
\usepackage[table]{xcolor}
\usepackage{siunitx}
\usepackage{rotating}
\usepackage{graphicx}
\usepackage{xcolor}
\usepackage{subcaption}
\usepackage{bbm}

\journal{}

\begin{document}

\begin{frontmatter}



\title{A Rank Graduation  metric for Algorithmic fairness} 


\author[1]{Dalia Atif}
\ead{atif.dalia@univ-tipaza.dz}

\author[2]{Paolo Giudici}
\ead{paolo.giudici@unipv.it}

\affiliation[1]{
    organization={Faculty of Economics, University of Tipaza},
    city={Tipaza},
    country={Algeria}
}

\affiliation[2]{
    organization={Department of Economics and Management, University of Pavia},
    city={Pavia},
    country={Italy}
}

\begin{abstract}
Fairness assessment in algorithmic decisions that affect individuals, such as credit scoring, often relies on parity measures calculated at the aggregate group level. Such measures may not reveal which individuals experience unfairness or which explanatory factors contribute to it. In this paper, we propose a rank-based framework that evaluates fairness through the distribution of model prediction errors, thereby linking fairness assessment with predictive accuracy and explainability. The framework combines Rank Graduation Fairness (RGF), its integrated measure AURGF, a centered Cram\'er--von Mises permutation test, and a feature removal procedure for fairness explainability.

We evaluate the methodology using logistic regression, random forest, gradient boosting, and a multilayer perceptron. The simulation study shows that protected-group imbalance can reverse descriptive fairness comparisons, whereas the proposed inferential procedure correctly distinguishes fair from unfair mechanisms. Its application to HMDA mortgage data produces model rankings that differ from those obtained with classical fairness criteria. Tree-based models, rather than logistic regression, provide the strongest combination of predictive accuracy and rank-based fairness, while the fairness null hypothesis is rejected for all four models. The persistence of disparity across statistical, bagging, boosting, and neural network specifications, together with the feature removal results, indicates that the observed unfairness is not specific to a single algorithm or predictor, but is associated with group differences embedded in the characteristics of the lending data. These findings support a broader approach to trustworthy artificial intelligence that combines predictive accuracy, fairness measurement, statistical inference, and explainability.
\end{abstract}

\begin{graphicalabstract}
\includegraphics[width=\textwidth]{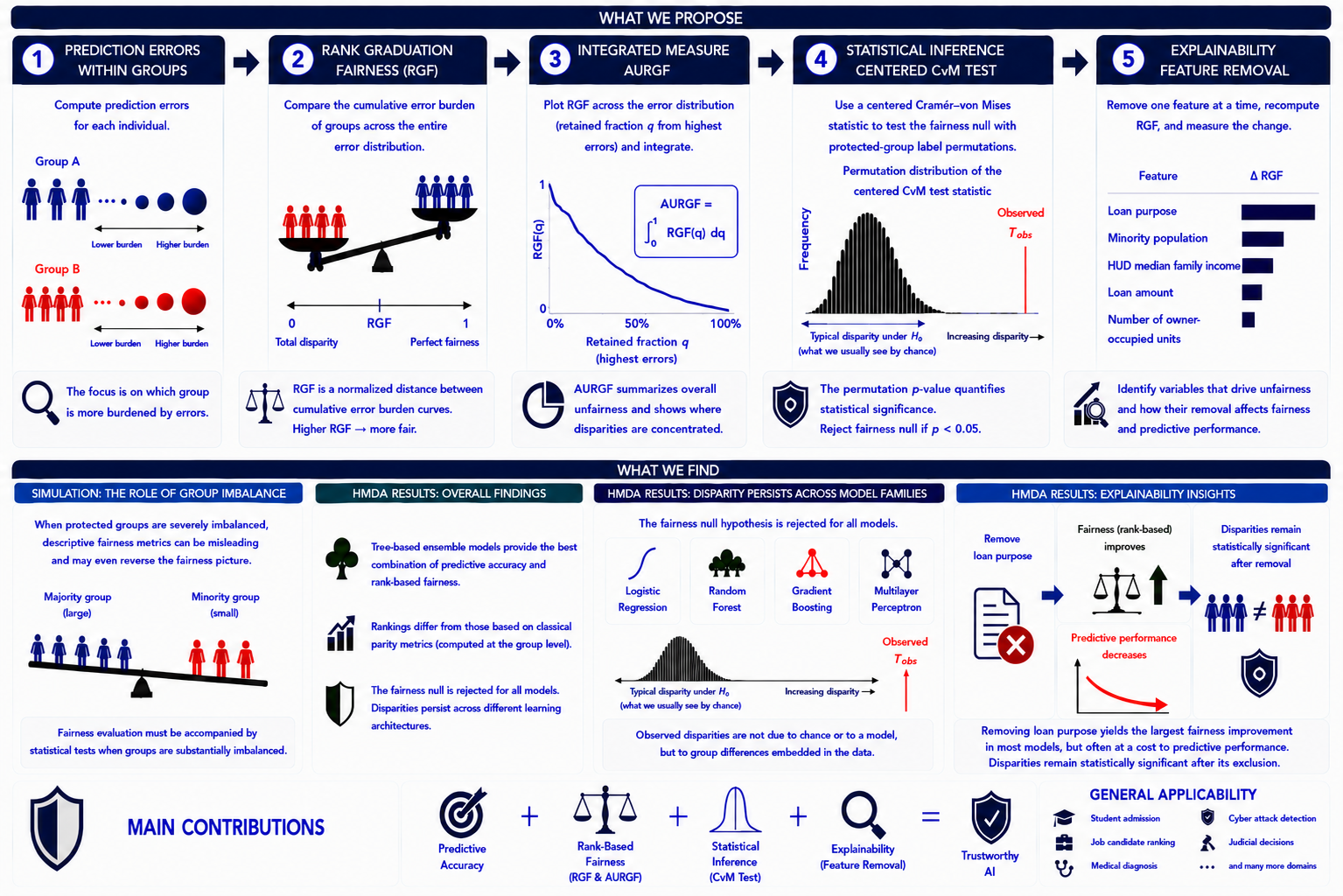}
\end{graphicalabstract}
\begin{highlights}

\item A Rank Graduation metric that extends SAFE AI to fairness assessment.

\item  The metric links fairness to ranked  prediction errors

\item  A Centered permutation test can correct for protected group imbalance.

\item The area under the metric can localize disparities across different individuals

\item The metric allows explainability of fairness disparities.

\item The methodology is applied to credit lending

\end{highlights}

\end{frontmatter}


\section{Introduction}
The use of machine learning in algorithmic decisions that affect individuals, such as credit lending, student admission, and job candidate ranking, has made governance a central issue in both academic research and regulatory practice. In this domain, predictive scores are not merely statistical outputs: they shape access to credit and the selection of students and job applicants, influence pricing and organizational procedures, and affect the distribution of opportunities across applicants. Evaluating such systems cannot be confined to predictive accuracy. It requires tools that examine how models behave, how their outputs can be explained, and whether their errors and decisions are distributed equitably across protected groups.

Explainable Artificial Intelligence (XAI) has been a major response to this governance challenge, especially in regulated fields such as credit lending, which we use as our reference.  In credit lending, explainability is required not only because complex classifiers may be opaque, but because model outputs must be open to scrutiny by institutions and supervisory authorities. The mainstream XAI literature has primarily focused on explaining predictions. Methods based on Shapley \citep{lundberg2017unified}, local surrogate models \citep{ribeiro2016should}, counterfactual explanations \citep{wachter2017counterfactual}, and feature removal procedures \citep{giudici2025safe} offer distinct approaches to identifying the variables that drive individual or aggregate model outputs. These tools have substantially improved black-box interpretability. However, they do not, by themselves, answer two central questions in regulated decision systems: which variables are associated with the emergence of unfairness? which individuals are most affected by unfairness?

This distinction is critical. A model may be explainable in the predictive sense while remaining opaque regarding its fairness properties. Conversely, a variable may have limited global predictive importance but still play a relevant role in generating unequal predictive reliability across protected groups. This concern is particularly relevant in credit lending, where recent studies emphasize both the need to identify variables that contribute to unfairness \citep{babaei2025explainability,hurlin2026fairness} and the importance of selecting fairness criteria appropriate for lending decisions \citep{kozodoi2022fairness}. In lending data, loan characteristics, income, collateral information, and neighborhood indicators may carry demographic or institutional information correlated with race or ethnicity. A model trained without protected attributes may therefore still generate disparities across protected groups through variables that operate as indirect channels of group differentiation.

The algorithmic fairness literature has provided a rich set of fairness metrics to evaluate disparities in automated decision systems. In credit scoring, these metrics are commonly organized around three major fairness criteria: independence, separation, and sufficiency \citep{kozodoi2022fairness}. Independence-based measures, such as demographic parity and disparate impact, compare the rates at which applicants from different protected groups receive favorable classifications. Separation-based measures, such as equal opportunity and equalized odds, compare classification error rates across groups conditional on the true outcome. Sufficiency-based measures, such as predictive parity, assess whether predicted positive classifications have comparable reliability across groups. These criteria remain essential because they are transparent, intuitive, and closely connected to regulatory discussions on discrimination in lending.
However, these metrics generally rely on a classification cutoff \citep{garg2020fairness}. Probabilistic scores are first converted into binary decisions, and fairness is then evaluated on the resulting predicted classes. Consequently, the fairness diagnosis does not assess the machine learning model alone, but the model combined with a specific decision rule. This makes fairness metrics sensitive to operational policy choices, since the selected cutoff may reflect institutional risk appetite, product design, market conditions, or supervisory requirements. A model may therefore appear more or less fair when the threshold changes, even if the underlying probabilistic score function remains unchanged.

This limitation matters in credit scoring because predicted probabilities are operationally meaningful before any final decision. They function as risk assessments and may influence review procedures, ranking, pricing, and internal monitoring. At the same time, credit scoring models are often estimated on accepted applicants, since repayment outcomes for rejected applicants are typically unavailable in the labeled modeling sample, creating a gap between the labeled training population and the full applicant population to which the score is applied \citep{ehrhardt2021reject}. By reducing probabilities to binary outcomes, threshold-based metrics further discard information about the severity of probabilistic misestimation. Two applicants may receive the same predicted class while being assigned substantially different denial probabilities. Similarly, two models may show comparable approval or denial disparities while concentrating severe probabilistic errors in different protected groups. A fairness audit based only on thresholded outcomes may therefore miss a relevant dimension of model behavior: the distribution and concentration of predictive unreliability across protected groups.

The SAFE AI literature provides a methodological basis for addressing this limitation. Its central insight is that AI governance should not rely on fragmented and incomparable indicators, but on metrics that are interpretable, auditable, and grounded in a common mathematical structure. Recent rank-graduation metrics operationalize this idea by evaluating accuracy, robustness, and explainability through ranked model outputs and controlled perturbations \citep{giudici2023safe, babaei2025rank}. Rank Graduation Accuracy measures the concordance between predictions and outcomes; Rank Graduation Robustness evaluates the stability of predictions under perturbation;  Rank Graduation Explainability assesses how model behavior changes when variables are removed. This framework matters not only for its individual metrics but also for its ability to treat different governance dimensions through consistent, rank-based logic.

This paper is positioned at the intersection of SAFE AI metrics, XAI, and computational fairness. It shifts the focus from explaining predictions to explaining fairness, introducing rank graduation fairness, which assesses whether predictive unreliability is unevenly distributed across protected groups and how this distribution changes under feature removal. The proposal is observational: it evaluates disparities using observed outcomes, model scores, and protected-group membership, without making counterfactual claims about discrimination \citep{castelnovo2022clarification}. By linking rank-based fairness measurement to explanations based on feature removal, this paper contributes to a more operational account of fairness in credit scoring models.

The remainder of the paper is structured as follows. Section~\ref{sec:literature} reviews the related literature. Section~\ref{sec:methodology} introduces the Rank Graduation Fairness proposal and its feature-removal extension.  
Section~\ref{sec:results} discusses the empirical findings obtained with both simulated and real data. Section~\ref{sec:conclusions} concludes.

\section{Literature Review}
\label{sec:literature}
\subsection{Fairness in Credit Scoring}
Recent research on fairness in credit scoring has increasingly moved from generic fairness definitions toward audit frameworks tailored to lending decisions. This shift reflects the specific nature of credit scoring, where models differentiate applicants by credit risk. At the same time, regulatory constraints require that such differentiation not result in unjustified disparities among protected groups. Within this literature, \cite{bono2021algorithmic} examine whether the transition from logistic scorecards to ensemble machine learning models alters the distributional properties of lending predictions. Using a large UK credit dataset, they show that machine learning models improve predictive accuracy without systematically worsening fairness across protected and sensitive subgroups. Their findings suggest that subgroup disparities are not necessarily a direct consequence of model complexity, but may also reflect credit data structure and the mechanics of risk prediction.
\cite{kozodoi2022fairness} offers a complementary perspective by evaluating fairness criteria and mitigation strategies in credit scoring models focused on profitability. Their analysis situates fairness interventions within the economic logic of lending, where measured group disparities, predictive performance, implementation constraints, and profitability interact. They identify separation as particularly relevant for credit scoring because it evaluates disparities in misclassification behavior under asymmetric costs of false approvals and false rejections. Fairness mitigation is therefore framed not as a purely statistical correction, but as a decision problem shaped by profitability and regulatory acceptability. Extending this literature toward formal diagnosis, \cite{hurlin2026fairness} extend the credit scoring fairness literature toward formal diagnosis and mitigation. They propose a statistical framework for testing fairness hypotheses and identifying variables associated with their rejection through Fairness Partial Dependence Plots. Their analysis is particularly relevant because it shows that simply removing a candidate variable and reestimating the model does not necessarily restore fairness and may induce substantial accuracy losses, partly because correlated variables can substitute for the removed feature. They therefore distinguish reestimation from feature neutralization, showing that muting selected variables can, in some cases, improve fairness with limited deterioration in predictive performance. Their contribution reflects a broader shift in the literature toward aligning fairness analysis with the specific statistical characteristics of lending decisions.

\subsection{Fairness and Explainability}
Recent studies use explainability as a diagnostic tool for fairness in credit scoring. \cite{agarwal2023countering} examine whether algorithmic lending models rely on applicant race when predicting loan rejection. Using HMDA mortgage data, they compare several machine learning models through Global Shapley Values and Shapley-Lorenz explanations. Their central contribution is to show that XAI can support ethical model selection. 
\cite{babaei2025explainability} extend this approach by showing that aggregate fairness diagnostics based on Shapley values may conceal conditional unfairness. Their analysis shows that race may appear to have limited influence in the full sample while becoming highly influential among applications for large loans, revealing a Simpson’s paradox effect. This result shows that fairness explanations should rely not only on global feature importance but also on whether protected-variable effects emerge within specific lending segments.  Applying this to HMDA data yields somewhat contradictory findings. While \cite{agarwal2023countering} obtain that Random Forest is more fair than logistic regression, \cite{babaei2025explainability} show that they are both unfair for high loan amounts, with Random Forest becoming fair at the aggregate level.  This indicates the need for further research to clarify for which individuals and features unfairness arises.

\subsection{SAFE AI Metrics and Rank-Based Governance}
The SAFE AI literature has established a quantitative framework for assessing the trustworthiness of artificial intelligence in finance. \cite{giudici2023safe} argue that financial AI systems should be evaluated beyond predictive accuracy, using statistical measures that capture sustainability, accuracy, fairness, and explainability. This contribution frames trustworthiness as an auditable property that can be monitored through quantitative indicators rather than as a purely regulatory principle.
\cite{giudici2025rga} provides the statistical foundation for this approach by introducing Rank Graduation Accuracy as a unified measure of predictive accuracy. RGA extends AUROC beyond binary classification to ordinal and continuous response variables by evaluating concordance between predicted and observed ranks.  Since RGA depends on prediction ranks, it is particularly well-suited to scoring systems and less sensitive to outlying observations than standard point-error measures.
Building on this rank-based logic, \cite{giudici2025safe} extend it into an integrated compliance framework. They propose three area under the curve metrics: AURGA, which measures how predictive accuracy deteriorates when ranked data is progressively removed; AURGR, which evaluates robustness by comparing prediction ranks before and after data perturbations; and AURGE, which measures explainability by assessing how strongly the ranking of model predictions changes relative to the full model. These metrics are then aggregated into a compliance score, enabling evaluation of accuracy, robustness, and explainability within a common mathematical framework. The main value of this framework is replacing fragmented model evaluation with a comparable auditing system that monitors multiple dimensions of artificial intelligence risk.

The present paper extends the rank-graduation architecture by developing fairness as a dedicated SAFE AI metric for algorithmic decisions, such as credit scoring. While existing SAFE AI metrics provide operational measures of accuracy, robustness, and explainability, fairness remains less formalized within the rank-graduation framework. 

We propose a Rank Graduation Fairness, which addresses this gap by evaluating where prediction errors concentrate across protected groups and whether the available features explain that concentration. In doing so, it moves fairness from a general compliance concern to an individual-level measurable dimension, aligned with the SAFE AI metric family.
\section{Methodology}
\label{sec:methodology}
\subsection{Research motivation and  contributions}

Fairness in machine learning can be understood both as a normative requirement and as a computational property of model behavior
\citep{pessach2023algorithmic}. In the normative sense, fairness expresses the expectation that algorithmic systems should not generate unjustified disadvantages for individuals or groups defined by protected characteristics. In the computational sense, fairness is translated into measurable restrictions on model outputs or rankings. This computational formulation is central to AI governance because it translates broad ethical and legal principles into quantifiable metrics that can be audited and incorporated into model validation procedures.

In algorithmic decisions, such as credit scoring, fairness assessment is commonly based on group-parity criteria, including demographic parity, disparate impact, equal opportunity, equalized odds, and predictive parity
\citep{hardt2016equality, barocas2016big, chouldechova2017fair,
kleinberg2016inherent}. These measures have advanced algorithmic fairness by providing transparent quantitative evidence of disparities across groups in both decisions and classification errors. They are especially useful when the policy concern is whether outcomes, such as loan approvals or denials, differ across protected groups. However, fairness metrics are not interchangeable. When protected groups have different base rates, satisfying one fairness criterion may render another mathematically unattainable or achievable only under restrictive conditions. Moreover, threshold choices and the practical observability of outcome statistics strongly affect which fairness metrics are meaningful in credit scoring applications \citep{garg2020fairness}.

The present paper addresses this issue by proposing a fairness framework that evaluates the distribution of prediction errors across protected groups. The central question is not only whether groups receive similar decisions, but also whether the burden of model error is distributed comparably across groups. This perspective is particularly relevant for policymakers because unfairness in algorithmic systems does not necessarily emerge only at the point of final decision. It may arise earlier through unequal reliability of risk scores, unequal exposure to severe prediction errors, or unequal concentration of predictive errors within particular populations.

The proposed framework is designed for models trained without using protected variables, such as gender, race, or ethnicity, as predictive inputs. The protected attribute is retained exclusively for fairness auditing. This design reflects a common regulatory setting in which protected characteristics should not be used as ordinary inputs to decision models \citep{zemel2013learning}, while remaining necessary for evaluating disparate effects. In such settings, excluding race from the predictive model does not, by itself, guarantee racial fairness. Other variables may encode demographic, geographic, socioeconomic, or institutional information associated with protected-group membership. A rigorous audit must therefore examine not only whether disparity exists, but also which remaining variables drive it.

This paper contributes to this problem by introducing a rank-based fairness and fairness explainability framework. The approach is consistent with the rank graduation logic developed in recent SAFE-AI research
\citep{giudici2025safe}. We extend this logic to the fairness domain by examining how removing each explanatory variable changes fairness itself. The focus therefore shifts from explaining predictions alone to explaining the sources of unequal predictive reliability.

The paper's first contribution is the definition of a Rank Graduation Fairness metric, denoted RGF. This metric evaluates whether prediction errors, measured as the distance between observed outcomes and predicted probabilities, are unevenly distributed across protected groups. Unlike classical parity measures that operate mainly on final binary decisions, RGF is computed directly from probabilistic model outputs. It therefore captures a distinct fairness dimension: the equality of predictive reliability across groups.

The second contribution concerns statistical testing. We use the
Cram\'er--von Mises (CvM) statistic to assess differences in the distribution of prediction errors across protected groups, with significance evaluated by permuting the protected-group labels. This avoids imposing a parametric distribution on the errors and allows the test to handle unequal group sizes. When the protected attribute contains more than two groups, inference is conducted through pairwise comparisons. The test therefore provides statistical support for the disparities identified by the rank-based fairness measures.

The third contribution is the Area Under the Rank Graduation Fairness Curve (AURGF). While RGF measures fairness across the full sample, AURGF assesses whether it is maintained across progressively larger portions of the ranked distribution of prediction errors. This distinction matters because an aggregate fairness score may conceal disparities concentrated among observations with the most severe prediction errors. By examining fairness over top burden fractions, AURGF characterizes both the magnitude and the location of disparities along the error distribution. From a policy perspective, this indicates whether unequal predictive reliability is concentrated among observations for which model errors are most consequential.

The fourth contribution extends rank-based fairness assessment to feature explainability. The proposed framework attributes observed fairness disparities to the predictive information carried by individual covariates, while retaining formal inference on the residual disparity through the Cram\'er--von Mises statistic and protected-group permutation testing. This provides an auditable basis for distinguishing features associated with the amplification or attenuation of predictive inequality, without assigning a causal interpretation to these associations.

From a guideline perspective, the proposed framework provides a unified basis for quantifying, localizing, and explaining disparities in predictive reliability. RGF enables direct comparison of rank-based fairness across competing models, while AURGF reveals where these disparities concentrate among observations. The permutation-based Cram\'er--von Mises procedure complements these measures with formal statistical inference and is designed to remain applicable under balanced or
imbalanced protected-group representation, as well as in settings involving multiple protected groups. Finally, the feature analysis identifies
covariates associated with the amplification or attenuation of observed
fairness disparities, thereby providing actionable evidence for model
governance and regulatory scrutiny.
\subsection{Rank Graduation Fairness}

 Building on the RGX principle of comparing Lorenz and concordance curves through normalized distances \citep{auricchio2026rank}, we define fairness as the absence of systematic group distortions in the ranking of prediction errors. In credit scoring, RGF evaluates whether predictive unreliability is disproportionately concentrated among protected groups before converting model scores into approval or denial decisions. This makes fairness a rank-based governance metric consistent with the broader SAFE AI architecture. Consider an evaluation sample:
\begin{equation}
    \mathcal{D}=\{(x_i,y_i,g_i)\}_{i=1}^{n}
\end{equation}
Where $x_i \in \mathbb{R}^{d}$ denotes the observed covariates, $y_i \in \{0,1\}$ is the realized outcome, and $g_i \in \mathcal{G}$ identifies the protected group of observation $i$. In the credit scoring application, $y_i=1$ denotes loan denial and $y_i=0$ denotes loan approval. The protected group variable is not included among the predictive covariates and is used only at the audit stage, after training the model. A probabilistic classifier $f$ is estimated using the admissible covariates and produces the score:
\begin{equation}
    \widehat{p}_i=f(x_i)=\widehat{\mathbb{P}}(Y_i=1\mid x_i)
\end{equation}
For each observation $i=1,\ldots,n$. The individual prediction error burden is denoted by:
\begin{equation}
    z_i = \ell(y_i,\widehat{p}_i)
\end{equation}
Where $\ell(\cdot,\cdot)$ is a non-negative loss function. This notation allows the framework to accommodate different specifications of predictive error, including the absolute error $\ell(y_i,\widehat{p}_i)=|y_i-\widehat{p}_i|$ and the squared error $\ell(y_i,\widehat{p}_i)=(y_i-\widehat{p}_i)^2$. Larger values of $z_i$ correspond to more severe probabilistic misestimation.

RGF is constructed from each protected group's cumulative contribution to the prediction error burden along the pooled error ranking. This common ranking underpins the specific group cumulative curves and their subsequent comparison. Let \(i_{(1)},\ldots,i_{(n)}\) denote the ordering of observations from the smallest to the largest prediction error burden:
\begin{equation}
z_{i_{(1)}} \leq z_{i_{(2)}} \leq \cdots \leq z_{i_{(n)}} 
\end{equation}

When error burdens are identical, treat the corresponding observations as a tie block. For each protected group, distribute the total burden within the block uniformly across its observations before constructing the cumulative group curves. This ensures that equal prediction-error burdens receive identical rank treatment and that the proposed fairness metrics are invariant to the ordering of tied observations. Using the pooled global ordering, the cumulative global error curve is defined at the empirical points \(t_k=k/n\) as:
\begin{equation}
L_z(t_k)
=
\frac{\sum_{j=1}^{k} z_{i_{(j)}}}{\sum_{i=1}^{n} z_i},
\qquad k=1,\ldots,n 
\end{equation}

The corresponding dual curve is obtained by cumulating the same error values in the opposite direction:
\begin{equation}
L_z^c(t_k)
=
\frac{\sum_{j=1}^{k} z_{i_{(n+1-j)}}}{\sum_{i=1}^{n} z_i},
\qquad k=1,\ldots,n 
\end{equation}

For each protected group \(g \in \mathcal{G}\), define the group error contribution at rank position \(j\) as:
\begin{equation}
z_{jg}
=
\mathbf{1}\{g_{i_{(j)}}=g\}z_{i_{(j)}} 
\end{equation}

Where \(\mathbf{1}\{g_{i_{(j)}}=g\}=1\) if the observation at rank position \(j\) belongs to group \(g\), and zero otherwise. The total error contribution of group \(g\), expressed relative to the total error of the full sample, is:
\begin{equation}
s_g
=
\frac{\sum_{j=1}^{n} z_{jg}}{\sum_{i=1}^{n}z_i}
\end{equation}

The raw cumulative error curve of group \(g\), evaluated along the pooled global ranking, is:
\begin{equation}
E_g(t_k)
=
\frac{\sum_{j=1}^{k} z_{jg}}{\sum_{i=1}^{n} z_i},
\qquad k=1,\ldots,n 
\end{equation}

Unlike a normalized group curve, \(E_g(t_k)\) is divided by the total error of all individuals. It therefore preserves group \(g\) 's contribution to the global error burden. However, this curve ends at \(s_g\), not at one. Directly comparing such curves across groups would therefore mix two different effects: the location of group errors in the pooled ranking and the total share of error carried by each group. To place all protected groups on a common cumulative scale, the remaining mass \(1-s_g\) is distributed uniformly over the pooled rank grid. The uniformly completed group curve is defined as:
\begin{equation}\label{eq:completed_curve}
\widetilde{E}_g(t_k)
=
E_g(t_k)
+
t_k(1-s_g),
\qquad k=1,\ldots,n 
\end{equation}

Equivalently,
\begin{equation}
\widetilde{E}_g(t_k)
=
\frac{\sum_{j=1}^{k} z_{jg}}{\sum_{i=1}^{n} z_i}
+
\frac{k}{n}
\left(
1-
\frac{\sum_{j=1}^{n} z_{jg}}{\sum_{i=1}^{n} z_i}
\right)
\end{equation}

This uniform completion is a neutral benchmark. The part of the total error that is not attributed to group \(g\) contains no group-specific rank information for that group. Allocating it uniformly over the rank grid avoids introducing an artificial concentration pattern. It also ensures that every group curve has the same boundary conditions:
\begin{equation}
\widetilde{E}_g(0)=0,
\qquad
\widetilde{E}_g(1)=1 
\end{equation}

The proposed metric is inspired by the RGX framework \citep{auricchio2026rank}, which expresses rank-based performance as a normalized absolute distance between cumulative curves. In this setting, we adapt this logic to fairness by comparing the uniformly completed cumulative error curves of protected groups. For two groups \(g\) and \(h\), the pairwise Rank Graduation Disparity is defined as:
\begin{equation}
RGD_{gh}
=
\frac{
\sum_{k=1}^{n}
\left|
\widetilde{E}_{g}(t_k)
-
\widetilde{E}_{h}(t_k)
\right|
}
{
\sum_{k=1}^{n}
\left|
L_{z}^{c}(t_k)-L_{z}(t_k)
\right|
}
\end{equation}

Equivalently, the corresponding pairwise Rank Graduation Fairness score can be written as:
\begin{equation}
RGF_{gh}
=
1
-
\frac{
\sum_{k=1}^{n}
\left|
\widetilde{E}_{g}(t_k)
-
\widetilde{E}_{h}(t_k)
\right|
}
{
\sum_{k=1}^{n}
\left|
L_{z}^{c}(t_k)-L_{z}(t_k)
\right|
}
\end{equation}

The numerator measures the absolute distance between the cumulative error profiles of groups \(g\) and \(h\) along the pooled error ranking. The denominator is the Gini-type benchmark associated with the global error distribution, obtained as the absolute distance between the cumulative error curve \(L_z(t)\) and its dual \(L_z^c(t)\). The resulting ratio therefore expresses group disparity relative to the total rank dispersion of prediction errors. This normalization guarantees a bounded metric, with values interpreted on a common scale between zero and one, provided that the pairwise curve distance does not exceed the Gini-type global dispersion benchmark.

Higher \(RGF\) values indicate that the two protected groups accumulate burden similarly along the pooled global ranking. Lower values indicate that the groups exhibit substantially different cumulative error-ranking patterns. For protected attributes with more than two categories, the metric is extended by averaging the pairwise Rank Graduation Disparities across all unordered group pairs:
\begin{equation}
RGD
=
\frac{2}{|\mathcal{G}|(|\mathcal{G}|-1)}
\sum_{g<h}
RGD_{gh}
\end{equation}

This construction differs from standard group fairness measures in three respects. First, it is computed on absolute probabilistic prediction errors rather than on thresholded classifications. Second, it compares the complete cumulative distributions of error burdens across groups using a common global ranking, rather than relying solely on average errors or decision rates. Third, it aligns with the Rank Graduation family of metrics, defining fairness as one minus the normalized absolute distance between ranked cumulative curves.

\subsubsection{Cram\'er--von Mises Permutation Test}

Statistical inference is conducted using a Cram\'er--von Mises (CvM) criterion. The inferential procedure uses the same uniformly completed group curves but explicitly accounts for the expected separation induced by unequal protected-group representation. For two groups \(g\) and \(h\), let:
\begin{equation}
    \pi_g=\frac{n_g}{n},
    \qquad
    \pi_h=\frac{n_h}{n}
\end{equation}
denote their sample proportions. Under the null hypothesis of exchangeability of labels, and conditional on the ordered error burdens, group \(g\) receives in expectation the proportion \(\pi_g\) of the cumulative error burden. Hence,
\begin{equation}
    \mathbb{E}_{H_0}
    \left[
        E_g(t_k)
    \right]
    =
    \pi_g L_z(t_k)
\end{equation}

From the definition of the completed group curve in
Equation~\eqref{eq:completed_curve} and noting that:
\begin{equation}
    \mathbb{E}_{H_0}[s_g]=\pi_g
\end{equation}
The expected completed curve is
\begin{equation}
    \mathbb{E}_{H_0}
    \left[
        \widetilde{E}_g(t_k)
    \right]
    =
    \pi_g L_z(t_k)
    +
    t_k(1-\pi_g)
\end{equation}

It follows that the expected difference between the completed curves of
groups \(g\) and \(h\) is:
\begin{equation}
    \mu_{gh}(t_k)
    =
    (\pi_g-\pi_h)
    \left[
        L_z(t_k)-t_k
    \right]
    \label{eq:expected_completed_difference}
\end{equation}

Equation~\eqref{eq:expected_completed_difference} is relevant when protected groups are unequally represented. Uniform completion imposes common endpoints on the group curves, but it does not imply identical expected trajectories under exchangeability. The expected difference vanishes when \(\pi_g=\pi_h\), whereas under group imbalance it may be non-zero whenever \(L_z(t_k)\neq t_k\). Accordingly, inference is based on the centered completed curve difference:
\begin{equation}
    D_{gh}^{c}(t_k)
    =
    \widetilde{E}_g(t_k)
    -
    \widetilde{E}_h(t_k)
    -
    \mu_{gh}(t_k),
\end{equation}
and the observed Cram\'er--von Mises statistic is defined as:
\begin{equation}
    T_{gh}^{obs}
    =
    \sum_{k=1}^{n}
    \left[
        D_{gh}^{c}(t_k)
    \right]^2
    \label{eq:centered_cvm}
\end{equation}

Thus, the test measures separation between the completed group curves beyond what exchangeability alone would predict from the observed group proportions. In the balanced case, \(\pi_g=\pi_h\) and therefore
\(\mu_{gh}(t_k)=0\), so that the centered statistic reduces to the squared
distance between the two completed curves.

Statistical significance is assessed by permutation rather than from the
standard Cram\'er--von Mises reference distribution. The two completed group curves are constructed from the same ranked prediction errors and are therefore dependent, making the usual independence-based CvM inference inappropriate in this setting \citep{curry2019rank}. We retain the CvM squared distance criterion and obtain its null distribution by permuting the protected-group labels while preserving the observed group sizes. For each permutation \(b=1,\ldots,B\), the completed curves are recomputed and the corresponding centered statistic \(T_{gh}^{(b)}\) is obtained. Following the finite-sample permutation \(p\)-value correction of \citet{phipson2016permutation}, the \(p\)-value is defined as:

\begin{equation}
    p_{gh}
    =
    \frac{
        1+
        \sum_{b=1}^{B}
        I\left(
            T_{gh}^{(b)}
            \geq
            T_{gh}^{obs}
        \right)
    }
    {B+1}
\end{equation}
Where \(I(\cdot)\) denotes the indicator function. A small \(p\)-value provides evidence against exchangeability, indicating that the observed separation in error accumulation is greater than would be expected from random group assignment given the observed group proportions. The theoretical basis of the permutation \(p\)-value is given in ~\ref{app:pvalue_theory}.

For a protected attribute comprising more than two groups, the same principle extends to the average pairwise centered statistic:
\begin{equation}
    T_{\mathcal{G}}^{obs}
    =
    \frac{2}
    {|\mathcal{G}|(|\mathcal{G}|-1)}
    \sum_{g<h}
    \sum_{k=1}^{n}
    \left[
        D_{gh}^{c}(t_k)
    \right]^2
    \label{eq:multigroup_cvm}
\end{equation}

For each permutation, labels are reassigned while preserving the observed sizes of all protected groups, and \(T_{\mathcal{G}}^{(b)}\) is recomputed. The corresponding multi-group permutation \(p\)-value is:
\begin{equation}
    p_{\mathcal{G}}
    =
    \frac{
        1+
        \sum_{b=1}^{B}
        I\left(
            T_{\mathcal{G}}^{(b)}
            \geq
            T_{\mathcal{G}}^{obs}
        \right)
    }
    {B+1}
\end{equation}

\begin{figure}[ht!]
    \centering
    \includegraphics[width=\textwidth]{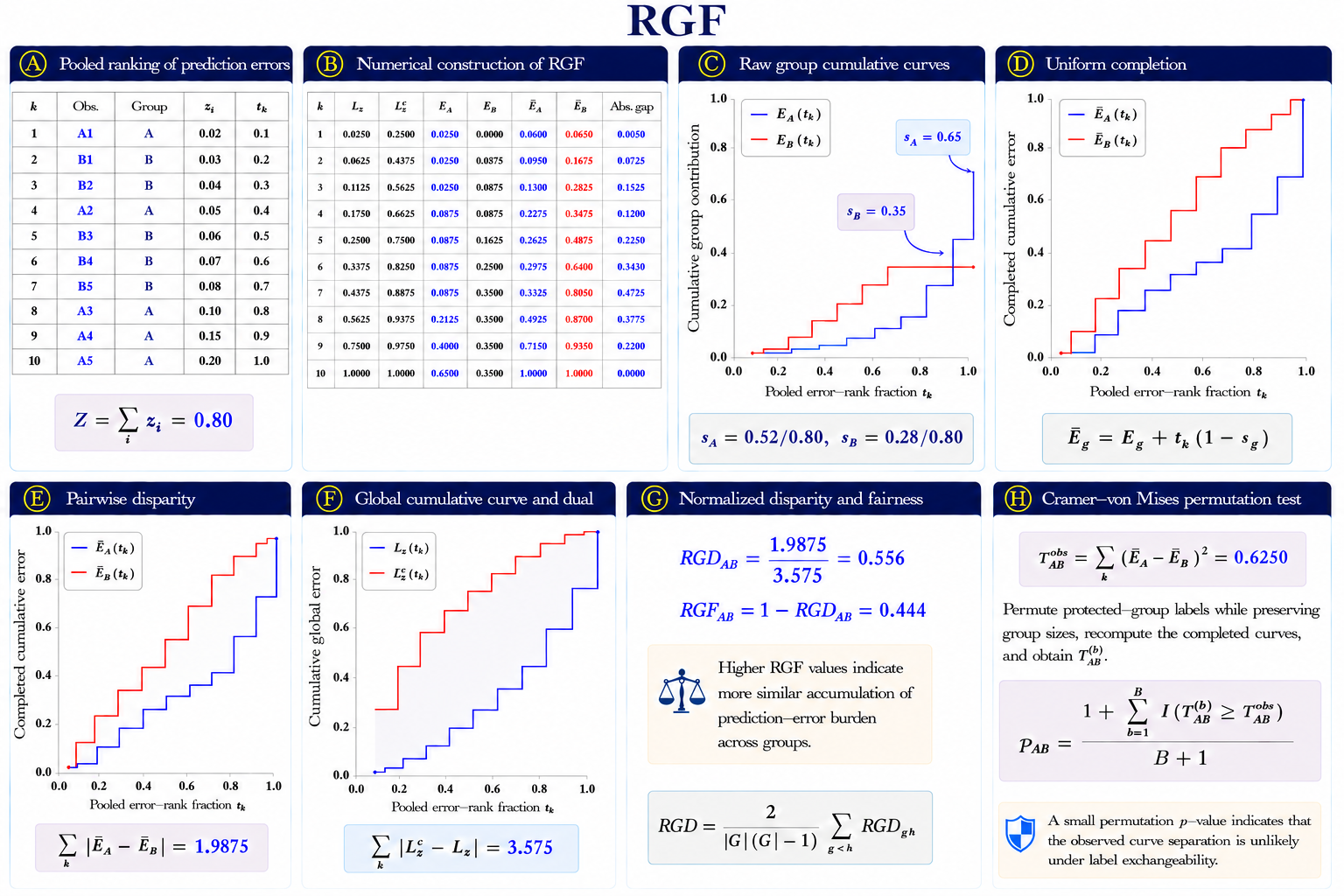}
    \caption{Illustration of the RGF construction: pooled error ranking and numerical construction (A--B), raw and uniformly completed group curves (C--D), pairwise disparity and normalized fairness measures (E--G), and Cram\'er--von Mises permutation inference (H).}
    \label{fig1}
\end{figure}

The resulting inference is therefore applicable to both balanced and
imbalanced protected-group configurations and extends naturally to
multi-group protected attributes. Figure~\ref{fig1} provides a numerical
illustration of the pooled error ranking, uniformly completed group curves,
RGD, RGF, and the associated permutation test.

\subsection{Area Under the Rank Graduation Fairness Curve}

The \(RGF\) metric summarizes fairness across the full sample. However, a single value may conceal disparities concentrated among the largest errors. To assess whether fairness holds across different severity levels, we introduce the Area Under the Rank Graduation Fairness Curve.

Inspired by AURGA \citep{giudici2025safe}, we evaluate fairness across progressively retained fractions of the ranked error distribution. For a retained fraction \(q\in(0,1]\), let \(D_q\) denote the subset containing the largest \(q\) fraction of error burdens. The Rank Graduation Fairness metric is recomputed within \(D_q\) using the same uniformly completed group cumulative curves defined above. Let \(RGD_{gh}(q)\) denote the pairwise Rank Graduation Disparity between groups \(g\) and \(h\) within \(D_q\):
\begin{equation}
RGD_{gh}(q)
=
\frac{
\sum_{k}
\left|
\widetilde{E}_{g,q}(t_k)
-
\widetilde{E}_{h,q}(t_k)
\right|
}
{
\sum_{k}
\left|
L_{z,q}^{c}(t_k)
-
L_{z,q}(t_k)
\right|
}
\end{equation}

Here, \(\widetilde{E}_{g,q}(t_k)\) and \(\widetilde{E}_{h,q}(t_k)\) are the uniformly completed cumulative error curves of groups \(g\) and \(h\) computed inside \(D_q\). The denominator represents the Gini-type benchmark for rank dispersion computed over the retained subset. For \(\mathcal{G}\) protected groups, the partial disparity is obtained by averaging all pairwise disparities:
\begin{equation}
RGD(q)
=
\frac{2}{|\mathcal{G}|\left(|\mathcal{G}|-1\right)}
\sum_{g<h}
RGD_{gh}(q)
\end{equation}

The corresponding partial fairness score is:
\begin{equation}
RGF(q)=1-RGD(q)
\end{equation}

\begin{figure}[ht!]
    \centering
    \includegraphics[width=1.0\textwidth]{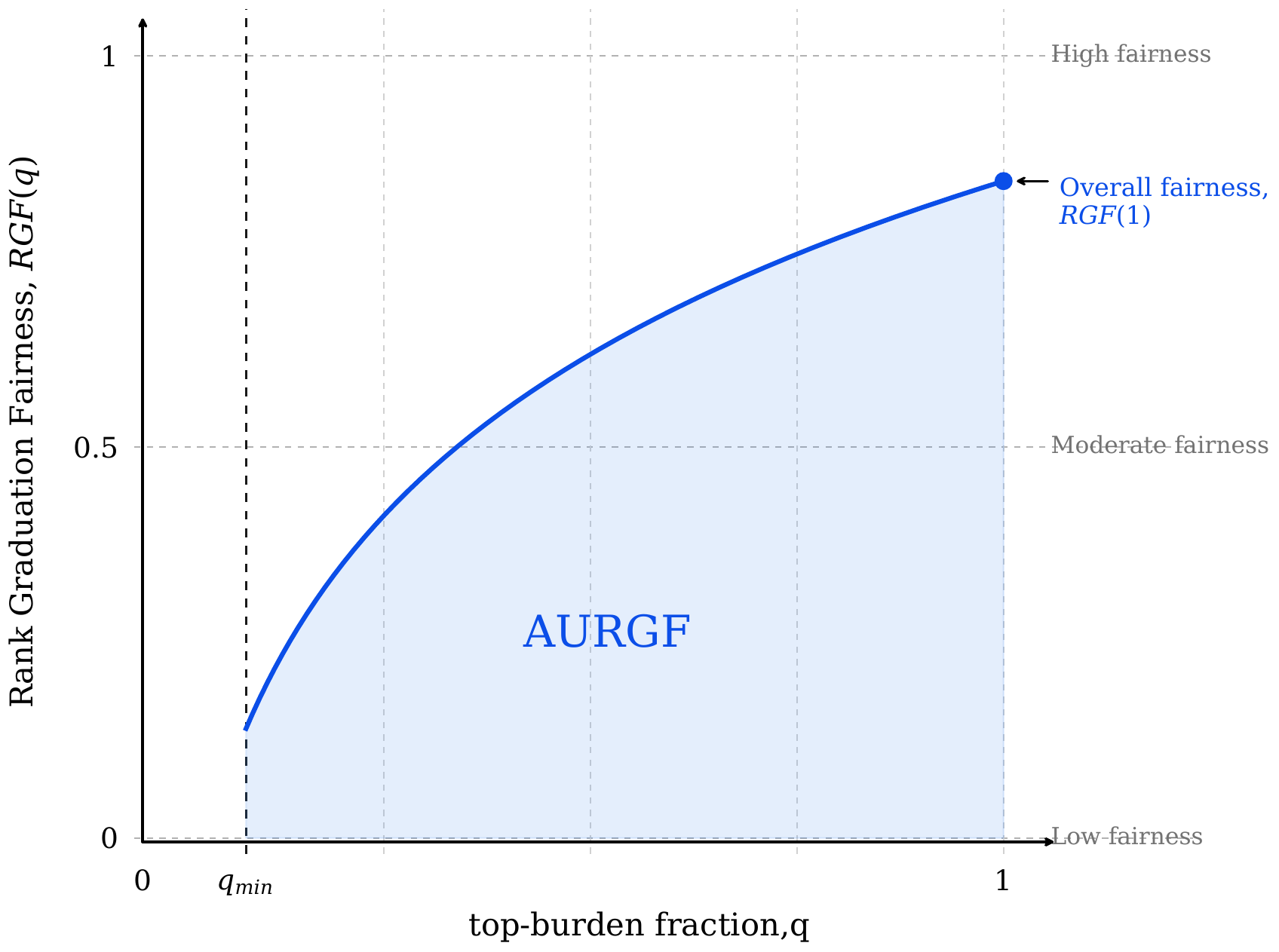}
    \caption{Conceptual representation of the Rank Graduation Fairness curve. The area under the curve summarizes fairness persistence across the ranked distribution of error burdens.}
    \label{fig2}
\end{figure}

The function \(RGF(q)\) defines the Rank Graduation Fairness Curve. As illustrated in Figure~\ref{fig2}, the curve shows how fairness evolves as the audit moves from the most severe errors to the full evaluation sample. The Area Under the Rank Graduation Fairness Curve is defined as:
\begin{equation}
AURGF
=
\frac{1}{1-q_{\min}}
\int_{q_{\min}}^{1} RGF(q)\,dq 
\end{equation}

In the empirical implementation, \(q_{\min}\) denotes the smallest retained fraction for which the protected-group comparison is well defined, meaning that each group included in the audit is represented in the retained subset and satisfies the minimum group-size requirement. For a grid \(q_1<q_2<\cdots<q_Q\), with \(q_1=q_{\min}\) and \(q_Q=1\), the integral is approximated by:
\begin{equation}
AURGF
\approx
\frac{1}{1-q_{\min}}
\sum_{\ell=1}^{Q-1}
\frac{
RGF(q_{\ell})+RGF(q_{\ell+1})
}{2}
(q_{\ell+1}-q_{\ell}) 
\end{equation}

A high \(AURGF\) indicates that predictive reliability remains comparable across protected groups over the ranked error distribution. A lower value indicates that fairness deteriorates in some regions of the distribution, especially among the largest prediction errors.

\subsection{Fairness Explainability Through Feature Removal}
\label{fair:explain}

The preceding metrics assess whether prediction error burdens vary across protected groups. A comprehensive fairness audit should also identify the admissible covariates that contribute to these disparities. This is particularly important when protected attributes are excluded from model training, as other variables may still capture demographic, geographic, socioeconomic, or institutional characteristics associated with group membership \citep{cornacchia2023auditing}. Let the admissible feature set be:
\begin{equation}
X=\{X_1,\ldots,X_{d_r}\}
\end{equation}

The full model, denoted by \(f_{\mathrm{full}}\), is estimated using all variables in \(X\). It produces predicted probabilities \(\widehat{p_i}\), error burdens \(z_i\), and the full-model fairness score \(RGF_{\mathrm{full}}\). The associated Cram\'er--von Mises permutation test gives the observed statistic \(T_{\mathrm{full}}^{obs}\) and the permutation \(p\)-value \(p_{\mathrm{full}}\). For each variable \(X_r\), the reduced feature set is:
\begin{equation}
X_{-r}=X\setminus\{X_r\}
\end{equation}

A reduced model \(f_{-r}\) is then estimated using \(X_{-r}\). It produces predicted probabilities:
\begin{equation}
\widehat{p}_{i,-r}=f_{-r}(x_{i,-r})
\end{equation}
and error burdens
\begin{equation}
z_{i,-r}=\ell(y_i,\widehat{p}_{i,-r})
\end{equation}

Using the reduced model errors \(z_{i,-r}\), the fairness score is recomputed and denoted by \(RGF_{-X_r}\). The Cram\'er--von Mises permutation test is also repeated, yielding the observed statistic \(T_{-X_r}^{obs}\) and the corresponding permutation \(p\)-value \(p_{-X_r}\). Here, \(T_{-X_r}^{obs}\) measures the curve separation after removing feature \(X_r\), while \(p_{-X_r}\) assesses whether the remaining group disparity is statistically significant. For each permutation \(b=1,\ldots,B\), protected-group labels are reassigned while preserving group sizes, and the statistic \(T_{-X_r}^{(b)}\) is recomputed. The permutation \(p\)-value is:
\begin{equation}
p_{-X_r}
=
\frac{
1+\sum_{b=1}^{B}
I\left(T_{-X_r}^{(b)}\geq T_{-X_r}^{obs}\right)
}
{B+1}
\end{equation}

The fairness effect of removing \(X_r\) is:
\begin{equation}
\Delta RGF_r
=
RGF_{-X_r}
-
RGF_{\mathrm{full}}
\end{equation}

A positive value of \(\Delta RGF_r\) indicates that removing \(X_r\) improves fairness. A negative value indicates that its removal worsens fairness. To compare effects across models, the normalized fairness contribution is defined as follows:

\begin{equation}
FC_r =
\begin{cases}
\displaystyle
\frac{RGF_{-X_r}-RGF_{\mathrm{full}}}
{1-RGF_{\mathrm{full}}},
& \Delta RGF_r \geq 0, \\[12pt]
\displaystyle
\frac{RGF_{-X_r}-RGF_{\mathrm{full}}}
{RGF_{\mathrm{full}}},
& \Delta RGF_r < 0 .
\end{cases}
\end{equation}

\(FC_r\) lies between \(-1\) and \(1\). Positive values suggest that \(X_r\) is associated with amplifying protected-group disparity. Negative values suggest that the variable may mitigate disparity. Values close to zero indicate limited influence on the fairness profile. The permutation test complements this descriptive measure. If removing \(X_r\) increases fairness and makes \(p_{-X_r}\) non-significant, the reduced model provides weaker evidence of group error concentration. If fairness decreases, or if \(p_{-X_r}\) remains significant, removing the feature does not eliminate the observed disparity.

This procedure extends feature-removal explainability to fairness auditing. The objective is not to infer a causal effect of \(X_r\), but to assess whether its predictive use is associated with a meaningful change in Rank Graduation Fairness and in the statistical evidence of group error concentration.

\section{Empirical findings}
\label{sec:results}

In this Section, we apply the proposed methodology to two distinct applications: a controlled simulated experiment and the well-known mortgage lending HMDA data.

\subsection{Simulated data}

We simulate a binary classification setting with $n=5{,}000$ observations and a deliberately imbalanced binary protected attribute. Let $R_i$ denote minority group membership. Group status is generated as:
\begin{equation}
    R_i \sim \operatorname{Bernoulli}(0.10)
\end{equation}
This implies that the expected proportions of majority and minority observations are $90\%$ and $10\%$, respectively. The protected attribute is generated independently of the covariates and does not enter either the outcome equation or the prediction models. This construction isolates the effect of unequal group representation from structural differences in observed characteristics or baseline risk.

Let $\operatorname{clip}_{[a,b]}(X)$ denote the restriction of a random variable $X$ to the interval $[a,b]$. Five continuous covariates are generated as follows:
\begin{align}
    \mathrm{Age}_i
    &=
    \operatorname{clip}_{[18,75]}
    \left\{
        \mathcal{N}(40,12^2)
    \right\}\\
    \mathrm{Income}_i
    &\sim
    \operatorname{Lognormal}(10.5,0.55^2)\\
    \mathrm{DebtRatio}_i
    &\sim
    \operatorname{Beta}(2.5,5)\\
    \mathrm{EmploymentYears}_i
    &=
    \operatorname{clip}_{[0,40]}
    \left\{
        \operatorname{Gamma}(3,3)
    \right\}\\
    \mathrm{CreditScore}_i
    &=
    \operatorname{clip}_{[300,850]}
    \left\{
        \mathcal{N}(650,70^2)
    \right\}
\end{align}

We also generate three categorical covariates as:
\begin{align}
    \mathrm{Education}_i
    &\sim
    \operatorname{Categorical}
    \left(
        0.45,\,
        0.40,\,
        0.15
    \right)\\
    \mathrm{EmploymentStatus}_i
    &\sim
    \operatorname{Categorical}
    \left(
        0.72,\,
        0.18,\,
        0.10
    \right)\\
    \mathrm{Urban}_i
    &\sim
    \operatorname{Bernoulli}(0.65)
\end{align}

The education levels correspond, in order, to secondary, bachelor, and postgraduate education. The employment status corresponds to employed, self-employed, and unemployed individuals. All covariates are mutually independent in the generating process and share the same distributions across protected groups. 

The binary response, indicating whether a loan request is accepted, is drawn from:
\begin{equation}
    Y_i \mid X_i
    \sim
    \operatorname{Bernoulli}(\pi_i),
    \qquad
    \pi_i
    =
    \operatorname{expit}(\eta_i)
\end{equation}
Where
\begin{align}
    \eta_i
    ={}&
    -1.10
    +0.015\,\mathrm{Age}_i
    +0.000015\,\mathrm{Income}_i
    -2.20\,\mathrm{DebtRatio}_i
    \nonumber\\
    &+
    0.025\,\mathrm{EmploymentYears}_i
    +0.006\,\mathrm{CreditScore}_i
    +0.30\,\mathbb{I}
    \left(
        \mathrm{Bachelor}_i
    \right)
    \nonumber\\
    &+
    0.60\,\mathbb{I}
    \left(
        \mathrm{Postgraduate}_i
    \right)
    -0.15\,\mathbb{I}
    \left(
        \mathrm{SelfEmployed}_i
    \right)
    \nonumber\\
    &-
    0.90\,\mathbb{I}
    \left(
        \mathrm{Unemployed}_i
    \right)
    +0.15\,\mathrm{Urban}_i
\end{align}

In the specification above, Secondary education and employed status are the reference categories. Note also that, as $R_i$ is absent from this specification, majority and minority observations follow the same conditional outcome process.

We split the data into $70\%$ training and $30\%$ test observations using joint stratification by protected-group membership and outcome status. This preserves both the protected-group imbalance and the class composition within each group. We standardize continuous variables, one-hot encode categorical variables, and estimate four machine learning classifiers: logistic regression, random forest, gradient boosting, and a multilayer perceptron. We retain the protected attribute exclusively for the subsequent fairness audit.

For each fitted classifier, we construct two evaluation settings from the same test sample. In the fair benchmark, the original predicted probabilities are retained:
\begin{equation}
    \widehat{p}_i^{\,F}
    =
    \widehat{p}_i
\end{equation}
Here, fairness refers to the absence of a group intervention rather than to exact equality of the realized group errors. This is because, under severe group imbalance, some finite-sample divergence remains possible even when the same generating mechanism governs both groups. 

The unfair setting introduces a controlled deterioration in the reliability of minority predictions. Let
\begin{equation}
    \widehat{\eta}_i
    =
    \operatorname{logit}
    \left(
        \widehat{p}_i
    \right)
\end{equation}
denotes the original predicted log-odds. We independently generate:
\begin{equation}
    \varepsilon_i
    \sim
    \mathcal{N}(0,2.20^2)
\end{equation}
and define
\begin{equation}
    \widehat{p}_i^{\,U}
    =
    \operatorname{expit}
    \left(
        \widehat{\eta}_i
        +
        R_i\varepsilon_i
    \right)
    \label{eq:unfair_prediction}
\end{equation}
Equation~\eqref{eq:unfair_prediction} leaves majority predictions unchanged, since $R_i=0$, while adding substantial logit-scale noise to minority predictions. This zero-mean disturbance increases the variability of minority predictions without imposing a directional shift on the logit scale. The construction holds the fitted model, covariates, outcomes, test observations, and majority predictions fixed across the two scenarios. The resulting difference can therefore be attributed specifically to a deterioration in minority group predictive reliability. We measure prediction burden using absolute probabilistic error.

\begin{figure*}[ht!]
\centering

\begin{subfigure}[t]{0.48\textwidth}
    \centering
    \includegraphics[width=\linewidth]{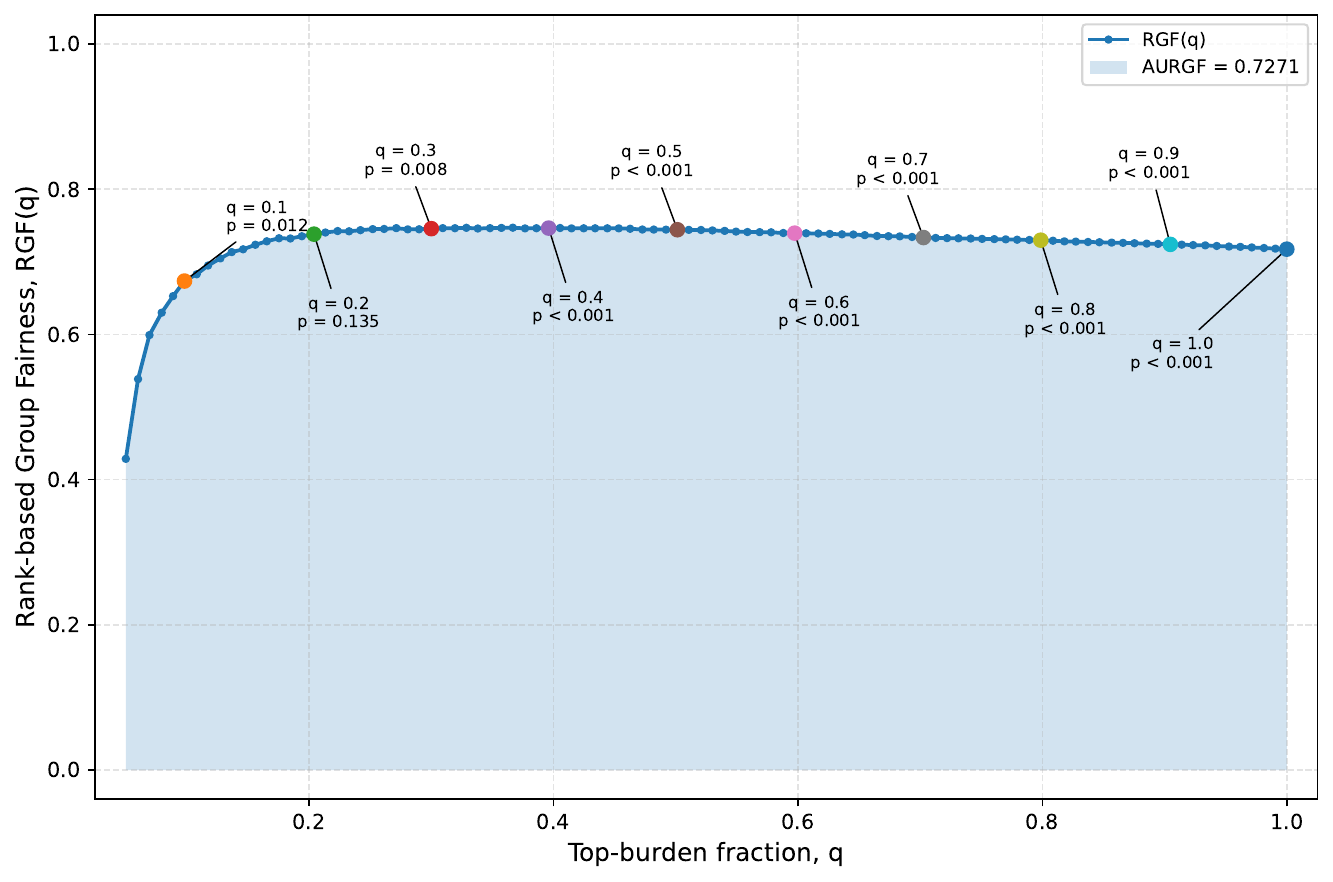}
    \caption{Logistic regression}
    \label{fig:aurgf_unfair_lr}
\end{subfigure}
\hfill
\begin{subfigure}[t]{0.48\textwidth}
    \centering
    \includegraphics[width=\linewidth]{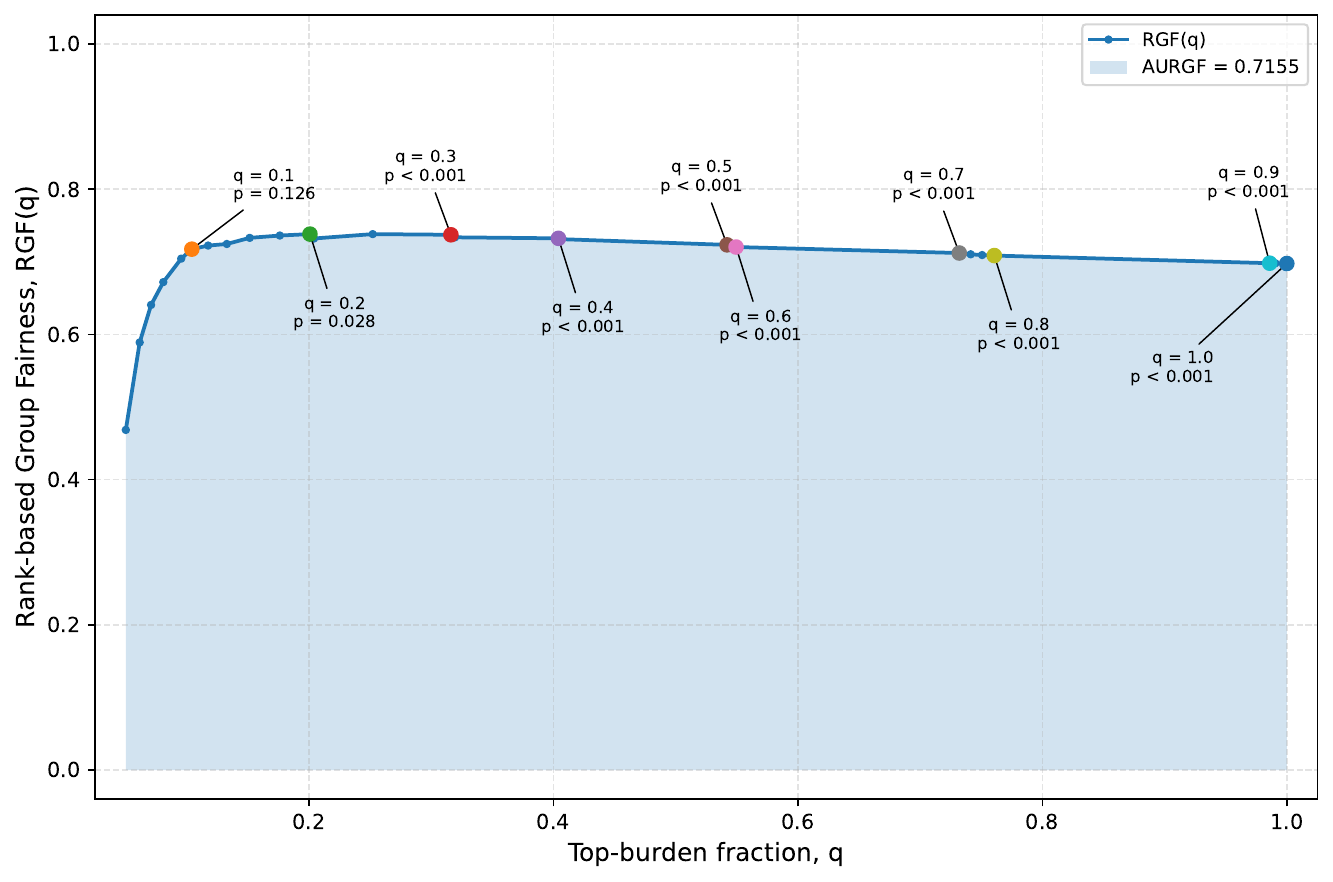}
    \caption{Random forest}
    \label{fig:aurgf_unfair_rf}
\end{subfigure}

\vspace{0.5cm}

\begin{subfigure}[t]{0.48\textwidth}
    \centering
    \includegraphics[width=\linewidth]{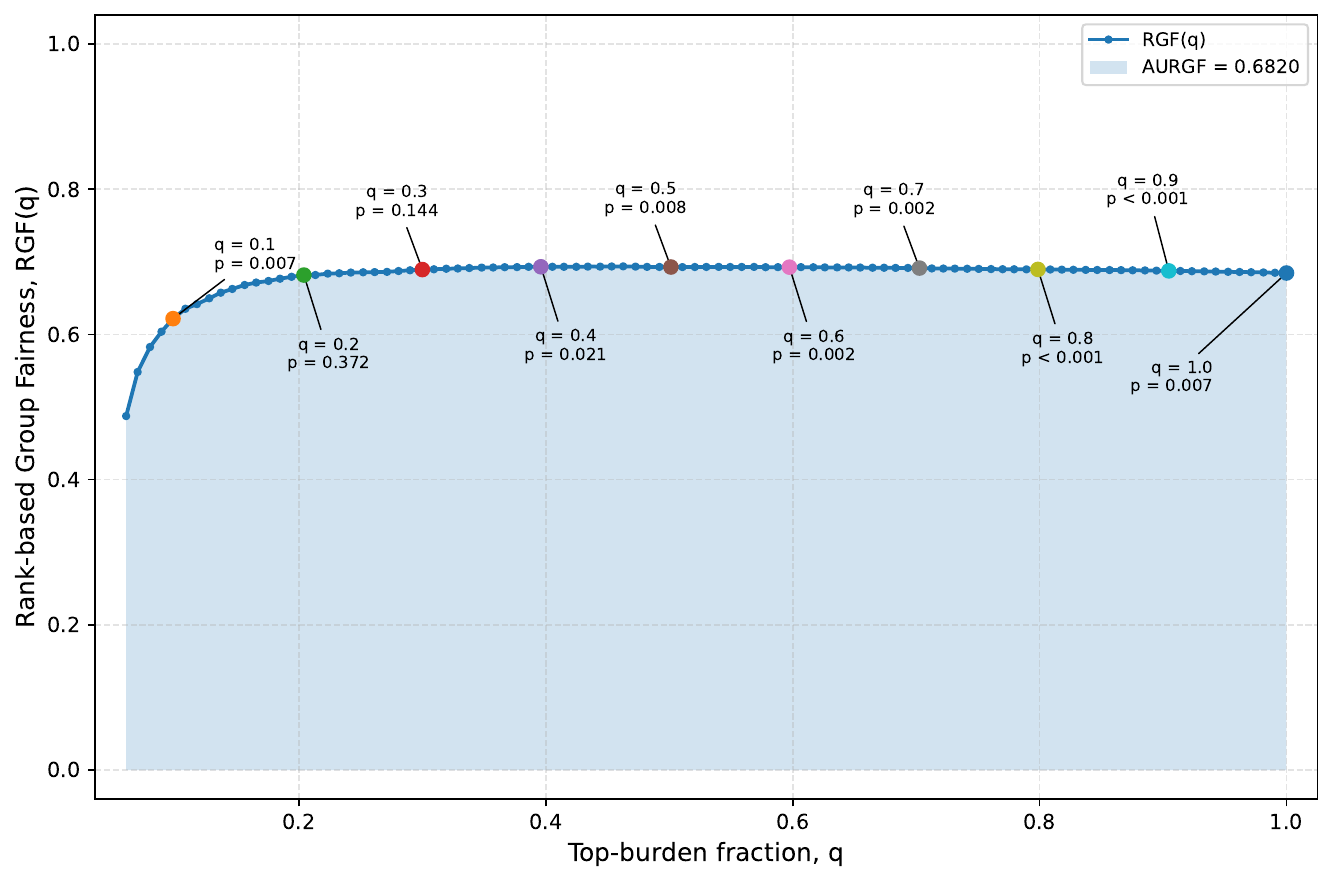}
    \caption{Gradient boosting}
    \label{fig:aurgf_unfair_gb}
\end{subfigure}
\hfill
\begin{subfigure}[t]{0.48\textwidth}
    \centering
    \includegraphics[width=\linewidth]{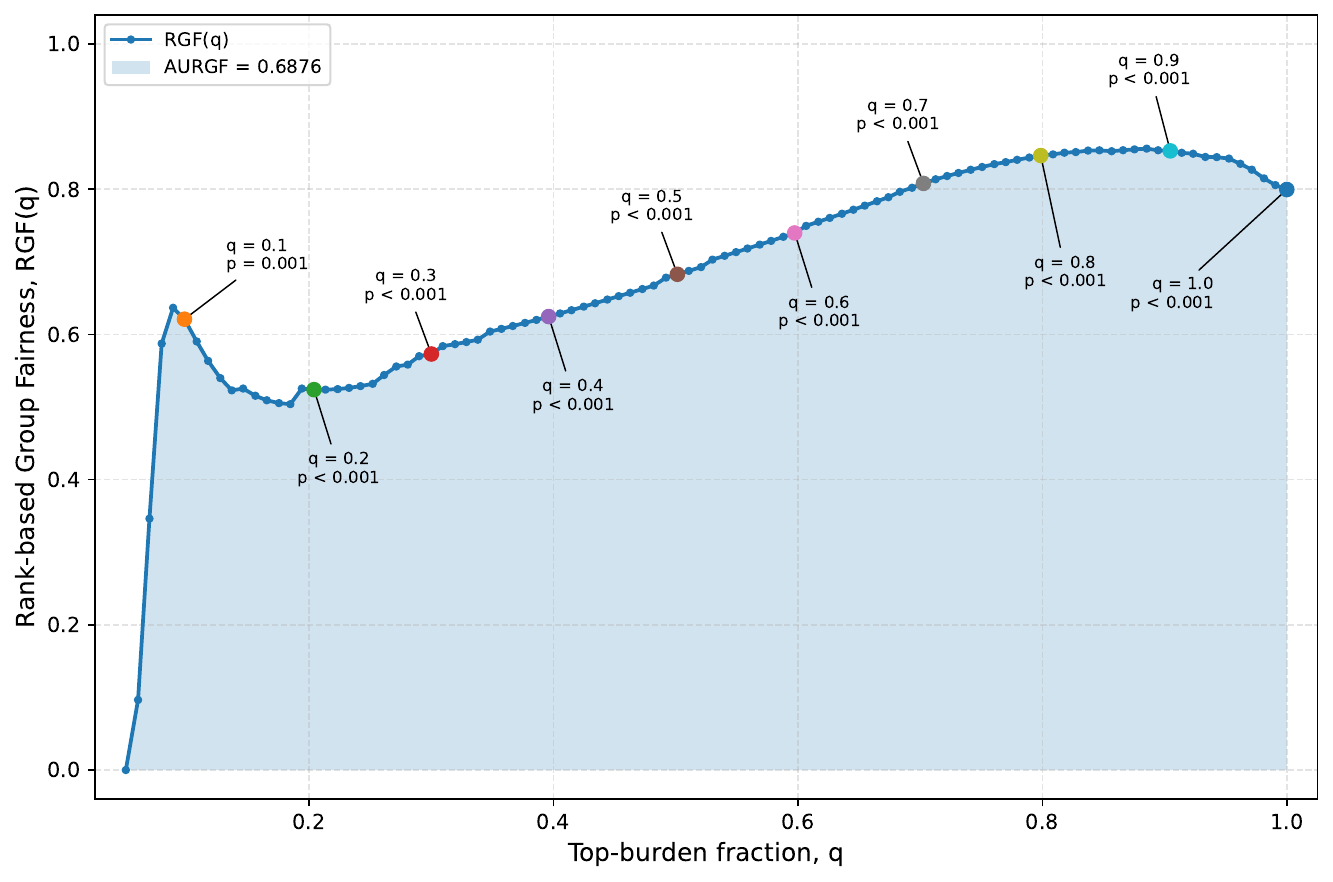}
    \caption{MLP}
    \label{fig:aurgf_unfair_mlp}
\end{subfigure}

\caption{
AURGF curves and associated permutation p-values for the four classifiers under the simulated unfair scenario. The horizontal axis represents the retained fraction \(q\) of observations with the largest error burdens. The shaded area corresponds to the Area Under the Rank Group Fairness curve (AURGF), while the annotations report permutation \(p\)-values at selected burden fractions. Smaller values of \(q\) correspond to the most severe prediction errors, with \(q=1\) indicating the full sample.
}
\label{fig:aurgf_unfair}
\end{figure*}

For each classifier and experimental scenario, we report RGD, RGF, and AURGF. Sampling uncertainty surrounding RGF is quantified through (2{,}000) group-stratified bootstrap replications, while statistical significance is evaluated using (2{,}000) protected-label permutations. Resampling is performed independently within each protected group, ensuring that the original (90{:}10) group composition is retained by construction. Table~\ref{tab:imbalanced-fairness-results} reports the complete simulation results. Complementing these aggregate estimates, Figure~\ref{fig:aurgf_unfair} traces the partial RGF curves underlying AURGF in the unfair scenario and reports the corresponding permutation (p)-values at selected burden fractions.

\begin{table*}[ht!]
\centering
\caption{Fairness diagnostics under protected-group imbalance}
\label{tab:imbalanced-fairness-results}

\begin{threeparttable}
\footnotesize
\setlength{\tabcolsep}{4.5pt}
\renewcommand{\arraystretch}{1.30}
\resizebox{0.97\textwidth}{!}{%
\begin{tabular}{
    @{}
    l
    l
    S[table-format=1.6]
    S[table-format=1.6]
    S[table-format=1.4]
    c
    S[table-format=1.4]
    S[table-format=2.4]
    S[table-format=1.4]
    @{}
}
\toprule

\multirow{2}{*}{\textbf{Model}} &
\multirow{2}{*}{\textbf{Scenario}} &
\multicolumn{2}{c}{\textbf{Group-level prediction error}} &
\multicolumn{3}{c}{\textbf{Rank-based fairness}} &
\multicolumn{2}{c}{\textbf{Statistical inference}} \\

\cmidrule(lr){3-4}
\cmidrule(lr){5-7}
\cmidrule(lr){8-9}

& &
{\makecell{\textbf{Majority}\\\textbf{MAE}}} &
{\makecell{\textbf{Minority}\\\textbf{MAE}}} &
{\textbf{RGD}} &
{\makecell{\textbf{RGF}\\{\scriptsize [95\% CI]}}} &
{\textbf{AURGF}} &
{\textbf{CvM-$T$}} &
{\textbf{$p$-value}} \\

\midrule

\multirow{2}{*}{Logistic regression}
& Fair
& 0.057022
& 0.051698
& 0.4138
& \makecell{0.5862\\[-1pt]{\scriptsize [0.5321, 0.6514]}}
& 0.5870
& 0.1560
& 0.6952 \\

\rowcolor{gray!8}
& Unfair
& 0.057022
& 0.110005
& 0.2825
& \makecell{0.7175\\[-1pt]{\scriptsize [0.6493, 0.7976]}}
& 0.7271
& 14.1567
& \multicolumn{1}{c}{\textbf{0.0009}} \\

\addlinespace[3pt]

\multirow{2}{*}{Random forest}
& Fair
& 0.058536
& 0.051959
& 0.4151
& \makecell{0.5849\\[-1pt]{\scriptsize [0.5389, 0.6409]}}
& 0.5810
& 0.2690
& 0.6177 \\

\rowcolor{gray!8}
& Unfair
& 0.058536
& 0.112069
& 0.3023
& \makecell{0.6977\\[-1pt]{\scriptsize [0.6373, 0.7684]}}
& 0.7155
& 12.3189
& \multicolumn{1}{c}{\textbf{0.0005}} \\

\addlinespace[3pt]

\multirow{2}{*}{Gradient boosting}
& Fair
& 0.053645
& 0.044183
& 0.4212
& \makecell{0.5788\\[-1pt]{\scriptsize [0.5280, 0.6427]}}
& 0.5790
& 0.5095
& 0.5287 \\

\rowcolor{gray!8}
& Unfair
& 0.053645
& 0.096404
& 0.3154
& \makecell{0.6846\\[-1pt]{\scriptsize [0.6227, 0.7595]}}
& 0.6820
& 9.3257
& \multicolumn{1}{c}{\textbf{0.0070}} \\

\addlinespace[3pt]

\multirow{2}{*}{Multilayer perceptron}
& Fair
& 0.284293
& 0.284516
& 0.3951
& \makecell{0.6049\\[-1pt]{\scriptsize [0.5412, 0.6719]}}
& 0.5717
& 0.0253
& 0.7946 \\

\rowcolor{gray!8}
& Unfair
& 0.284293
& 0.369214
& 0.2004
& \makecell{0.7996\\[-1pt]{\scriptsize [0.7681, 0.8212]}}
& 0.6876
& 10.1641
& \multicolumn{1}{c}{\textbf{0.0005}} \\

\bottomrule
\end{tabular}
}
\vspace{2mm}

\begin{minipage}{0.94\textwidth}
\scriptsize
\textit{Notes:}
MAE denotes the mean absolute probabilistic prediction error,
while CvM--$T$ denotes the centered observed Cramér--von Mises statistic.
Values in brackets are 95\% confidence intervals for RGF, obtained from
group-stratified bootstrap replications. Permutation $p$-values are
based on protected-label permutations; values significant at the
5\% level are shown in bold. Shaded rows identify the unfair scenario.
\end{minipage}

\end{threeparttable}
\end{table*}
Table~\ref{tab:imbalanced-fairness-results} reveals an important consequence of severe protected-group imbalance. Across all four classifiers, RGF is higher in the unfair scenario than in the fair benchmark, contrary to the direction implied by the simulation design. This reversal reflects the descriptive RGF measure's sensitivity to the highly unequal group representation in this experiment. It should therefore not be interpreted as evidence that the induced deterioration in minority predictions improves fairness.

The inferential results lead to the opposite, and intended, conclusion. The
centered Cram\'er--von Mises statistic explicitly removes the completed curve separation expected under exchangeability at the observed group proportions. Accordingly, the permutation tests do not reject the null in the fair scenario, whereas they reject it for every classifier after perturbing minority group predictions. The simulation thus separates two issues: the finite-sample behavior of the descriptive RGF score under severe group imbalance and the detection of systematic distributional disparity once statistical inference accounts for the induced imbalanced component. This behavior is not unique to the proposed framework: established group fairness measures have also been shown to be sensitive to protected-group proportions \citep{brzezinski2024properties}, reinforcing the distinction between a descriptive fairness estimate and the statistical evidence supporting a disparity.

Figure~\ref{fig:aurgf_unfair} traces the evolution of RGF over progressively larger fractions of the highest error burdens and reports the
corresponding AURGF values. As with the full-sample RGF results, AURGF reverses under severe protected-group imbalance, yielding comparatively high values despite the deliberate deterioration imposed on minority group predictions. The permutation results nevertheless recover the distinction introduced by the simulation through the centered CvM statistic.

The point at which the null hypothesis is rejected differs across classifiers. For gradient boosting, significance emerges only after retaining a larger fraction of the ranked observations. This result is consistent with its error profile in the unfair scenario: gradient boosting records the lowest minority MAE among the four classifiers (\(0.0964\)), indicating that the imposed perturbation translates into a comparatively smaller overall error burden for this model. The group separation is consequently less pronounced among the most extreme errors and becomes detectable only as the retained fraction expands. This should not be interpreted as an intrinsic requirement of gradient boosting for a larger sample, but as a feature of the model's error distribution. The MLP provides a useful contrast. Its substantially larger prediction errors under the unfair scenario produce a detectable group departure at smaller retained fractions, despite relatively high RGF values at the upper end of the error ranking.

The random forest curve is additionally affected by the occurrence of tie
blocks. Repeated predicted probabilities generate identical absolute error
burdens for multiple observations, so a requested retained fraction that
intersects a tie cannot be implemented by arbitrarily splitting the tied
observations. The entire block is retained instead, causing the realized
fraction \(q\) to move to the end of the block. The resulting jumps in the
random forest curve therefore reflect the tie-preserving construction of the partial RGF rather than changes in the underlying definition of the metric. This treatment ensures that RGF and AURGF are invariant to the arbitrary ordering of observations with identical error burdens.

\subsection{HMDA data}

To illustrate the proposed methodology, we use data collected under the Home Mortgage Disclosure Act (HMDA), the most comprehensive publicly available source of information on the U.S. mortgage market. Enacted in 1975 and implemented through Regulation C, HMDA requires covered financial institutions to report detailed information on individual mortgage applications. These data help assess whether lenders are adequately serving their communities' credit needs, inform the development of equitable housing policies, and help identify potential patterns of discriminatory lending. To protect applicants’ privacy, the publicly released records are anonymized. Our empirical analysis focuses on 446,902 mortgage applications filed in New York State in 2017. Following \cite{giudici2025safe}, we prepare the data by excluding irrelevant or non-informative variables and consolidating infrequent categories where appropriate.

 {\sloppy
We preserve the existing imbalance in the protected variable, applicant\_race\_1. The dataset comprises two racial groups, with 90.63\% in the majority group and 9.37\% in the minority group. The binary target variable indicates whether a loan application is declined. It is named action\_taken and is imbalanced, with approximately 18.37\% of loans declined.
\par} 

 {\sloppy
The analysis includes five categorical and eight continuous covariates. The categorical covariates are loan\_type, applicant\_sex, co\_applicant\_sex, loan\_purpose, and lien\_status. The continuous covariates are loan\_amount\_000s, applicant\_income\_000s, population, minority\_population, hud\_median\_family\_income, tract\_to\_msamd\_income, number\_of\_owner\_occupied\_units, and number\_of\_1\_to\_4\_family\_units. 
\par} 

The data reflects the default and the protected-group imbalances commonly encountered in lending portfolios. It thus provides a realistic benchmark for assessing fairness under realistic representation. The dataset is partitioned into training and test subsets using a 70\%--30\% split with joint stratification on action\_taken and
applicant\_race\_1. This strategy simultaneously preserves the outcome distribution and the protected-group proportions in both subsets. We standardize continuous predictors and one-hot encode categorical predictors. We exclude the protected variable from model estimation and use it only during the post hoc fairness audit.

We have applied four different machine learning models to the data. Logistic regression was tuned with \(C=82.4431\), a convergence
tolerance of \(1.24\times10^{-5}\), the \texttt{saga} solver, and no
class weighting. A random forest was then fit, with 700 trees and
bootstrap sampling, a log-loss splitting criterion, a maximum depth of
12, a maximum-feature proportion of \(0.5\), a minimum leaf size of 8,
a minimum split size of 5, and balanced class weighting. Gradient
boosting was then specified with 256 trees, a learning rate of
\(0.1802\), a maximum depth of 5, a maximum-feature proportion of
\(0.5\), a subsampling rate of \(0.9308\), a minimum leaf size of 14,
and a minimum split size of 10. Finally, we fitted an MLP with a single hidden layer containing \(128\) neurons with ReLU activation, a batch size of 512, an \(L_2\) regularization parameter of \(1.54\times10^{-3}\), an
initial learning rate of \(2.11\times10^{-3}\), and early stopping based on a no-improvement window of 10 iterations.

\begin{figure*}[t]
\centering

\begin{subfigure}[t]{0.48\textwidth}
    \centering
    \includegraphics[width=\linewidth]{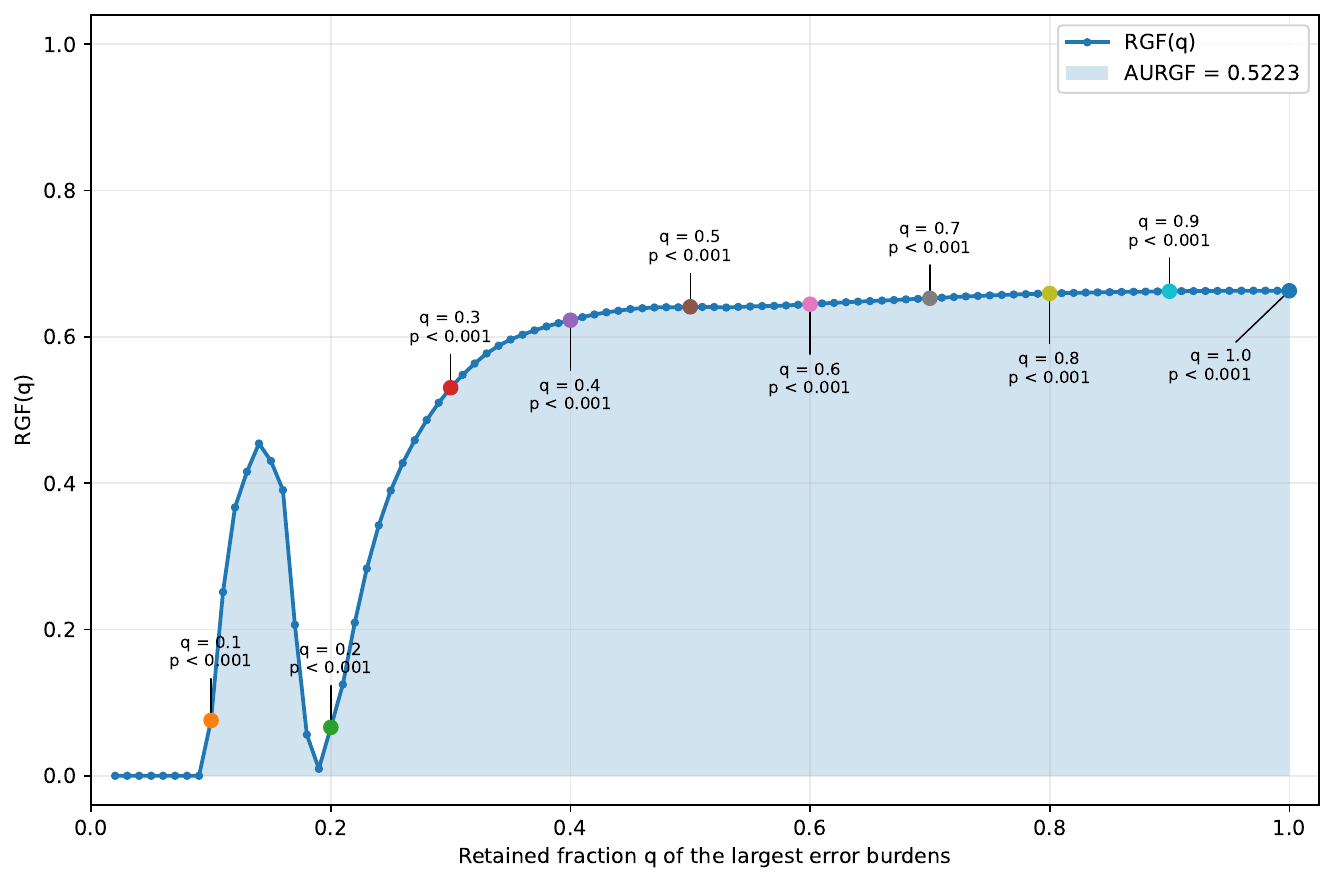}
    \caption{Logistic regression}
    \label{fig:aurgf_lr}
\end{subfigure}
\hfill
\begin{subfigure}[t]{0.48\textwidth}
    \centering
    \includegraphics[width=\linewidth]{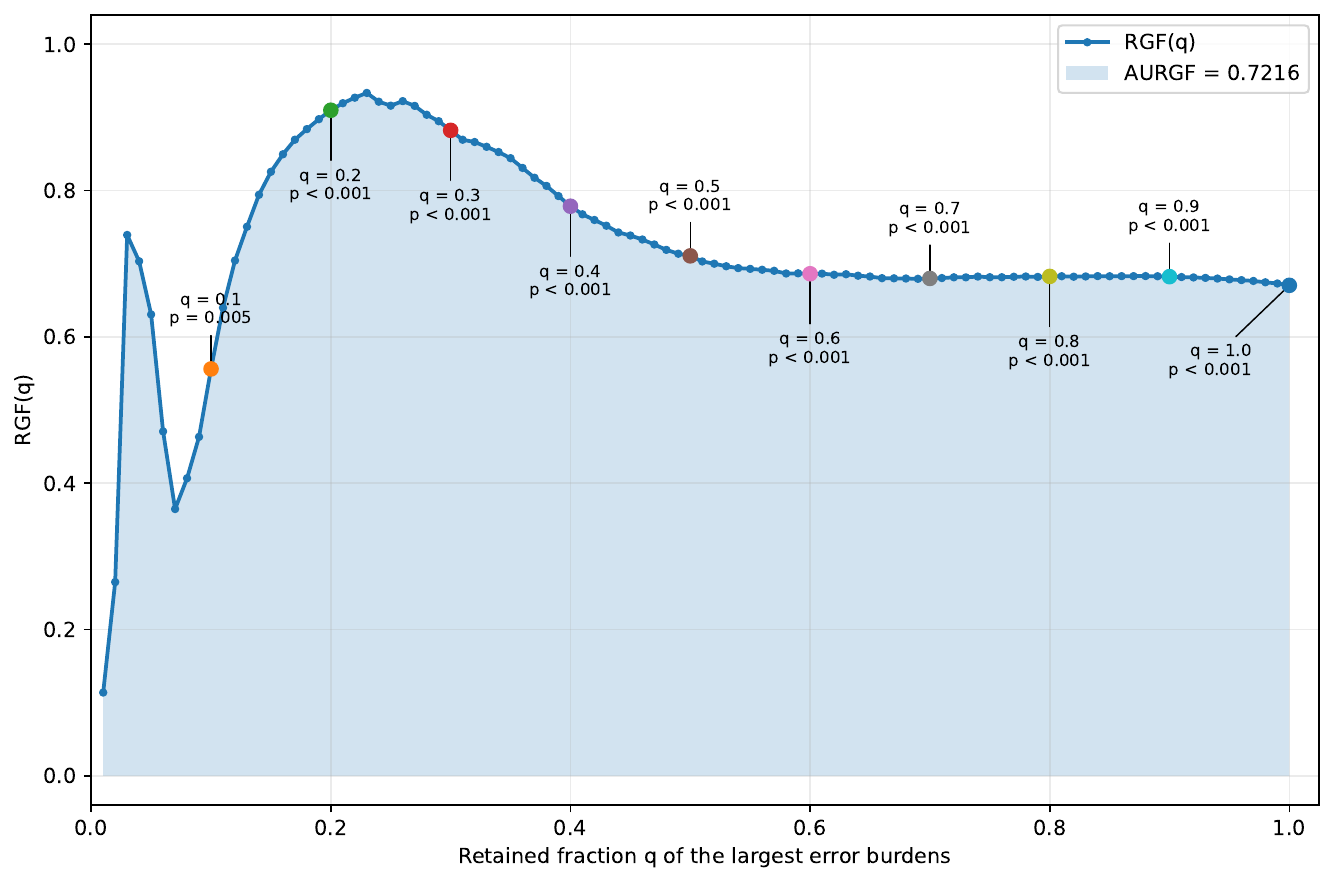}
    \caption{Random forest}
    \label{fig:aurgf_rf}
\end{subfigure}

\vspace{0.5cm}

\begin{subfigure}[t]{0.48\textwidth}
    \centering
    \includegraphics[width=\linewidth]{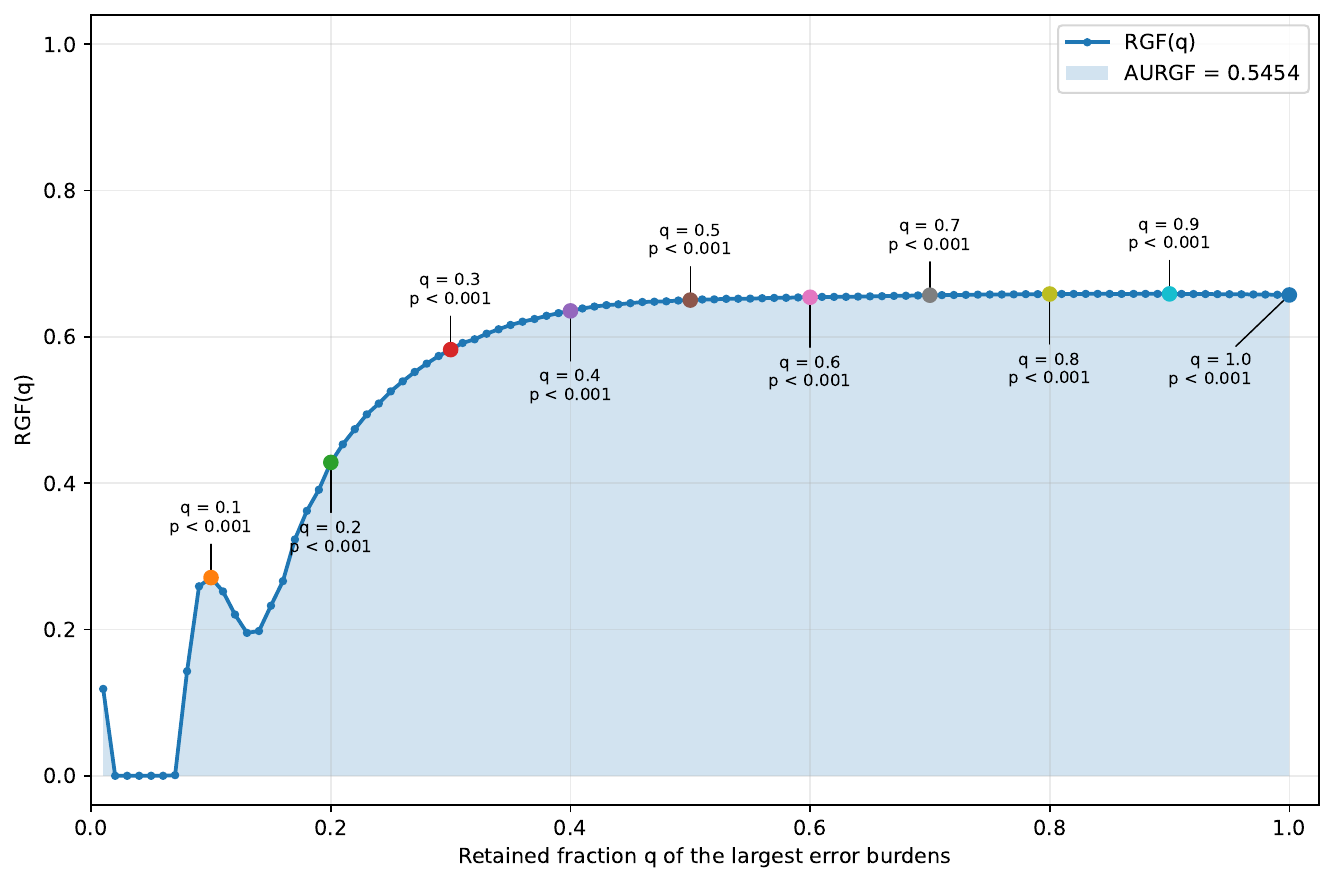}
    \caption{Gradient boosting}
    \label{fig:aurgf_gb}
\end{subfigure}
\hfill
\begin{subfigure}[t]{0.48\textwidth}
    \centering
    \includegraphics[width=\linewidth]{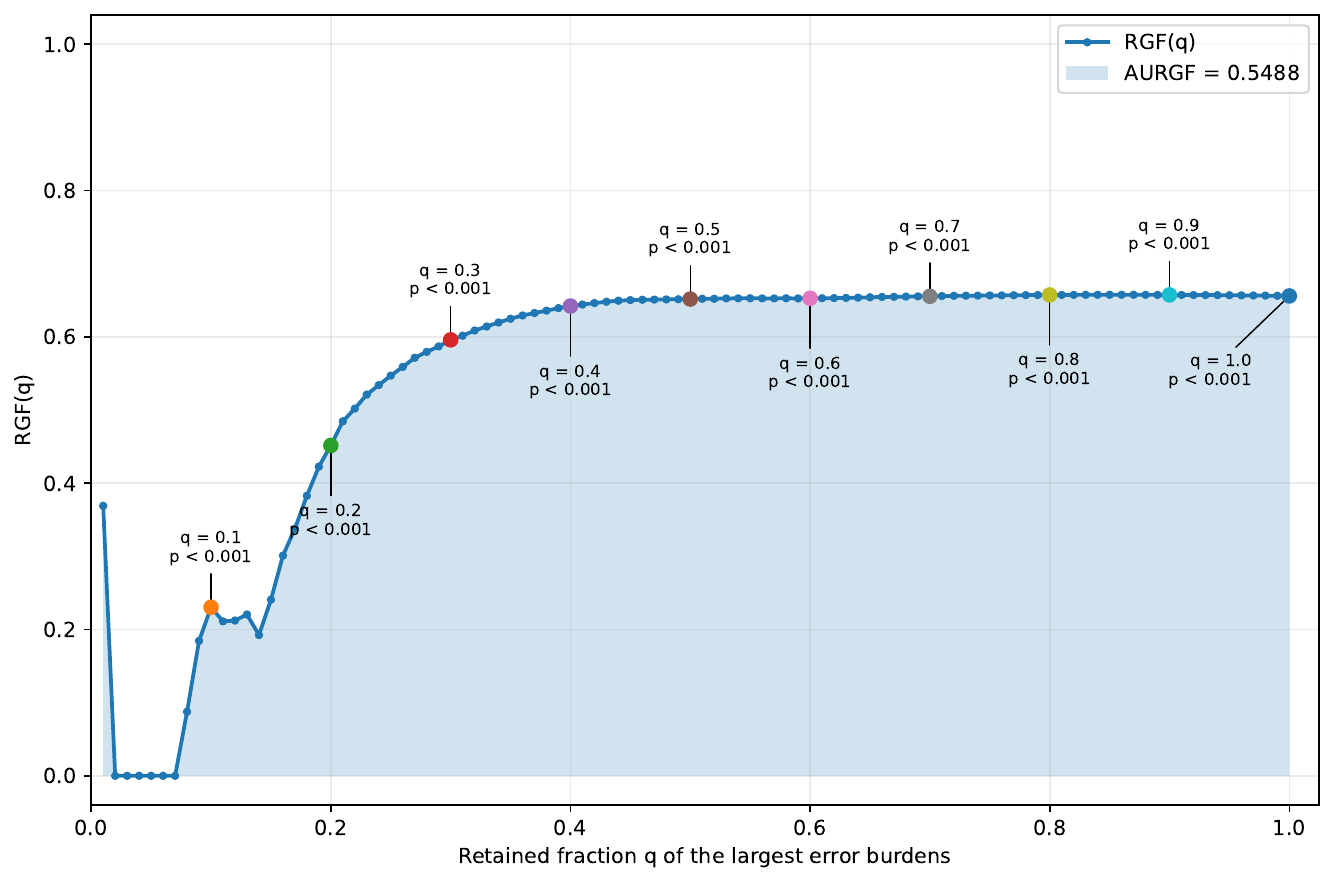}
    \caption{MLP}
    \label{fig:aurgf_mlp}
\end{subfigure}

\caption{
AURGF curves for the four classifiers. The horizontal axis represents the retained fraction \(q\) of the observations with the largest error burdens. The shaded area corresponds to the Area Under the Rank Group Fairness curve (AURGF). Labels indicate permutation \(p\)-values computed at selected retained fractions. Smaller \(q\) values focus on the highest-error observations, whereas \(q=1\) corresponds to the full test sample.
}
\label{fig:aurgf_curves}
\end{figure*}

The models were compared on a held-out test partition which
preserves the original protected-group imbalance. Throughout the fairness audit, we measure prediction burden using the absolute probabilistic error, \(|y_i-\widehat{p}_i|\), consistent with the proposed RGF framework. The resulting predictive performance and fairness estimates are reported in Table~\ref{tab:imbalanced_model_comparison}, whereas the corresponding AURGF curves together with permutation \(p\)-values are presented in Figure~\ref{fig:aurgf_curves}.

We quantify the statistical uncertainty associated with the proposed fairness measures using 2,000 group-stratified bootstrap replications to construct 95\% confidence intervals for RGF. We assess statistical significance using 2,000 protected-group label permutations, and report the centered Cramér--von Mises statistic to assess differences
in the distribution of prediction errors across protected groups.

\begin{table*}[ht!]
\centering
\caption{Predictive performance and rank-based fairness results on the original dataset with imbalanced protected groups}
\label{tab:imbalanced_model_comparison}

\begin{threeparttable}
\small
\setlength{\tabcolsep}{5pt}
\renewcommand{\arraystretch}{1.15}
\resizebox{0.97\textwidth}{!}{%
\begin{tabular}{
    l
    cc
    cc
    c
    c
    c
    c
    c
}
\toprule
& \multicolumn{2}{c}{Predictive performance}
& \multicolumn{4}{c}{Rank-based fairness}
& \multicolumn{2}{c}{Statistical inference}
& \\

\cmidrule(lr){2-3}
\cmidrule(lr){4-7}
\cmidrule(lr){8-9}

Model
& ROC--AUC
& PR--AUC
& RGD
& RGF
& \makecell{RGF\\95\% CI}
& AURGF
& CvM--$T$
& Permutation \(p\)-value
& \(q_{\min}\) \\
\midrule

Logistic regression
& 0.6794
& 0.3198
& 0.3370
& 0.6630
& [0.6580, 0.6684]
& 0.5223
& 78.5705
& \textbf{0.0005}
& 0.0200 \\

Random forest
& 0.7348
& 0.4213
& \textbf{0.3295}
& \textbf{0.6705}
& [0.6637, 0.6787]
& \textbf{0.7216}
& \textbf{19.4222}
& \textbf{0.0005}
& 0.0100 \\

Gradient boosting
& \textbf{0.7364}
& \textbf{0.4232}
& 0.3427
& 0.6573
& [0.6523, 0.6628]
& 0.5454
& 72.7159
& \textbf{0.0005}
& 0.0100 \\

MLP
& 0.7303
& 0.4108
& 0.3441
& 0.6559
& [0.6510, 0.6615]
& 0.5488
& 70.2581
& \textbf{0.0005}
& 0.0100 \\

\bottomrule
\end{tabular}
}
\vspace{2mm}

\begin{minipage}{0.94\textwidth}
\scriptsize
\textit{Notes:}
Results are computed on the original dataset, preserving the natural imbalance between the protected groups. Values in brackets are 95\% group-stratified bootstrap confidence intervals for RGF. CvM--\(T\) denotes the centered observed Cramér--von Mises statistic, with the bold value indicating the smallest statistic.
Permutation \(p\)-values are based on protected-label permutations. Bold entries in the predictive-performance and AURGF columns indicate the best result, whereas the bold RGD and RGF values correspond to the smallest disparity and highest fairness, respectively.
Bold \(p\)-values denote statistical significance at the 5\% level.
The quantity \(q_{\min}\) denotes the smallest top-burden fraction used to
construct the partial RGF curve.
\end{minipage}

\end{threeparttable}
\end{table*}

Table~\ref{tab:imbalanced_model_comparison} shows that Gradient Boosting and Random Forest achieve the strongest predictive performance, with Gradient Boosting attaining the highest ROC--AUC and PR--AUC. From a fairness perspective, Random Forest records the highest AURGF and the lowest centered Cramér--von Mises statistic. The latter indicates that Random Forest exhibits the smallest departure between the prediction error distributions of the two protected groups among the models considered; it should not, however, be interpreted as evidence of fairness. Indeed, the permutation test rejects the null hypothesis for Random Forest (\(p=0.0005\)), as it does for all other
classifiers.

This distinction matters given the pronounced imbalance in protected-group representation. The simulation results show that severe imbalance can affect the ordering of the descriptive RGF and AURGF measures, and these quantities should therefore be interpreted cautiously in this setting. The permutation inference, based on the centered Cramér--von Mises statistic, accounts for the separation expected from the observed group proportions. Although Random Forest presents the smallest distributional discrepancy in relative terms, the evidence remains sufficient to reject fairness. More generally, the empirical results distinguish relative fairness across models from statistical evidence of fairness: a model may compare favorably with competing specifications while still exhibit a significant disparity across protected groups.

Figure~\ref{fig:aurgf_curves} further examines these results by tracing RGF over progressively larger fractions of the highest prediction error burdens and summarizing this trajectory through AURGF. Across all four models, RGF is lowest among the most severe prediction errors and increases as less extreme observations are included. This supports the intuition that unfairness tends to be concentrated where model predictions are least accurate. Given the pronounced protected-group imbalance, however, interpret RGF and AURGF cautiously and consider them jointly with the inferential evidence.

From a credit lending perspective, Table~\ref{tab:imbalanced_model_comparison} provides consistent evidence of unfairness across all four models considered for the HMDA data. This finding is in line with \cite{babaei2025explainability},
but differs from the conclusions reported by \cite{agarwal2023countering}. An important distinction is that the evidence reported here is supported by formal permutation inference, rather than by descriptive fairness measures alone. Rejecting the fairness null across all model specifications therefore indicates that the observed disparities cannot be attributed solely to the magnitude or ordering of the descriptive fairness scores.
\subsubsection{HMDA data, fairness explainability}
\label{explain}
We now consider the empirical implementation of the fairness
explainability framework developed in Section~\ref{fair:explain}. The analysis examines each predictor's contribution to each classifier's rank-based fairness under the original imbalanced protected-group representation. Specifically, each predictor is removed in turn, the
corresponding model is re-estimated, and the resulting change in RGF is used to quantify its normalized fairness contribution, \(FC_r\). This procedure identifies whether a given predictor contributes to or mitigates disparities in the distribution of error burdens across protected groups.

\begin{figure*}[t]
\centering

\begin{subfigure}[t]{0.48\textwidth}
    \centering
    \includegraphics[width=\linewidth]{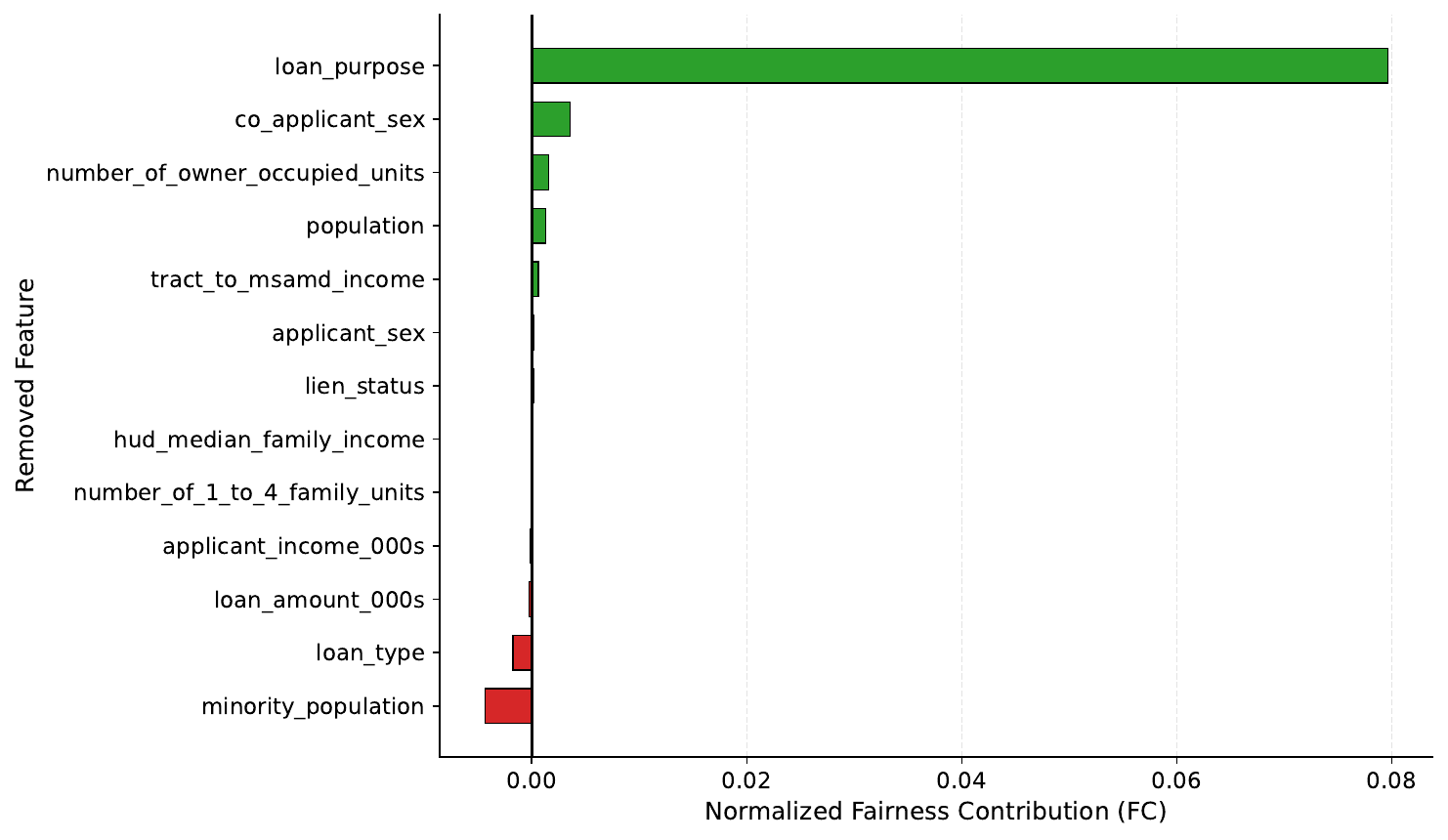}
    \caption{Logistic regression}
    \label{fig:explainability_lr}
\end{subfigure}
\hfill
\begin{subfigure}[t]{0.48\textwidth}
    \centering
    \includegraphics[width=\linewidth]{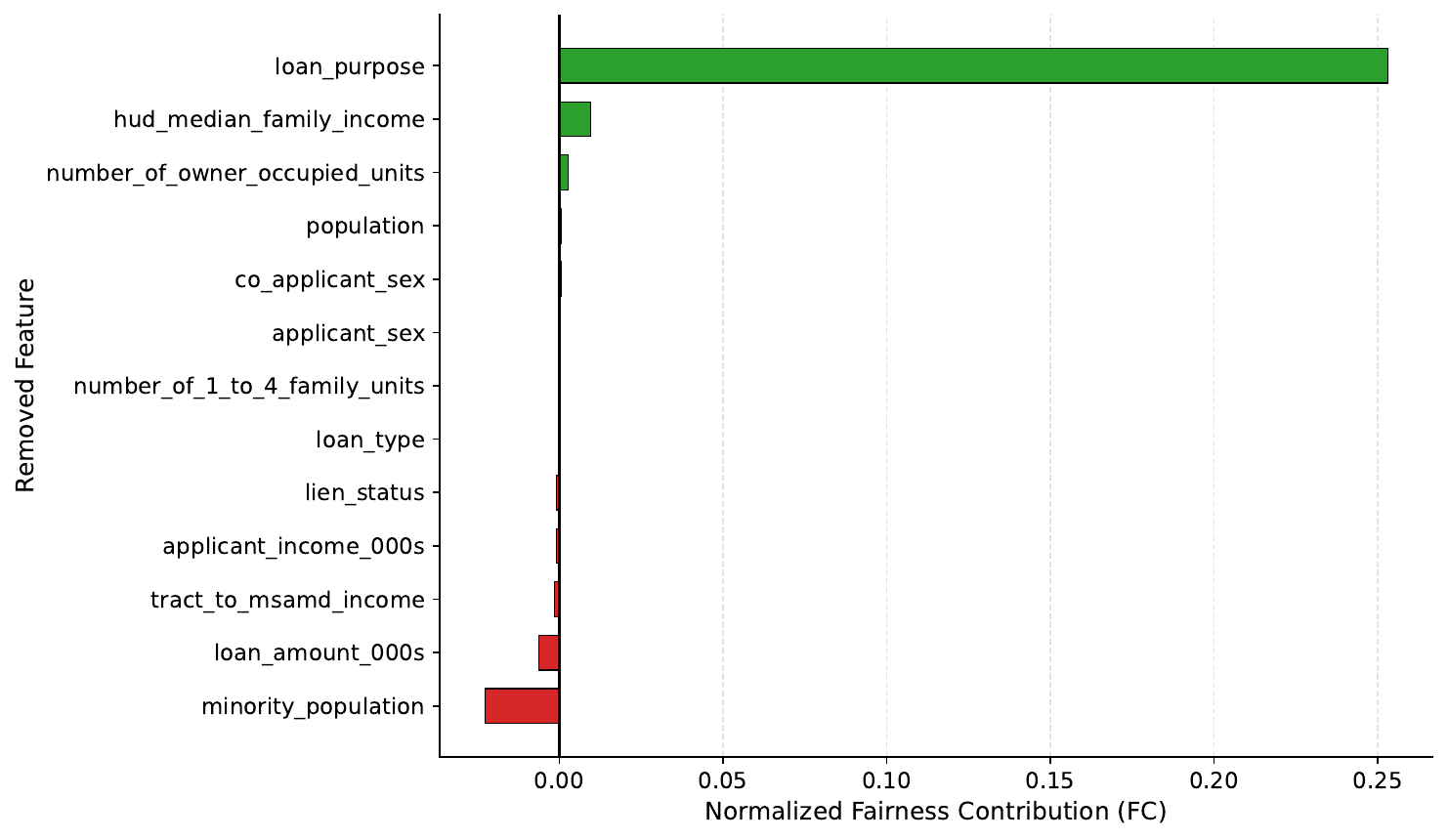}
    \caption{Random forest}
    \label{fig:explainability_rf}
\end{subfigure}

\vspace{0.5cm}

\begin{subfigure}[t]{0.48\textwidth}
    \centering
    \includegraphics[width=\linewidth]{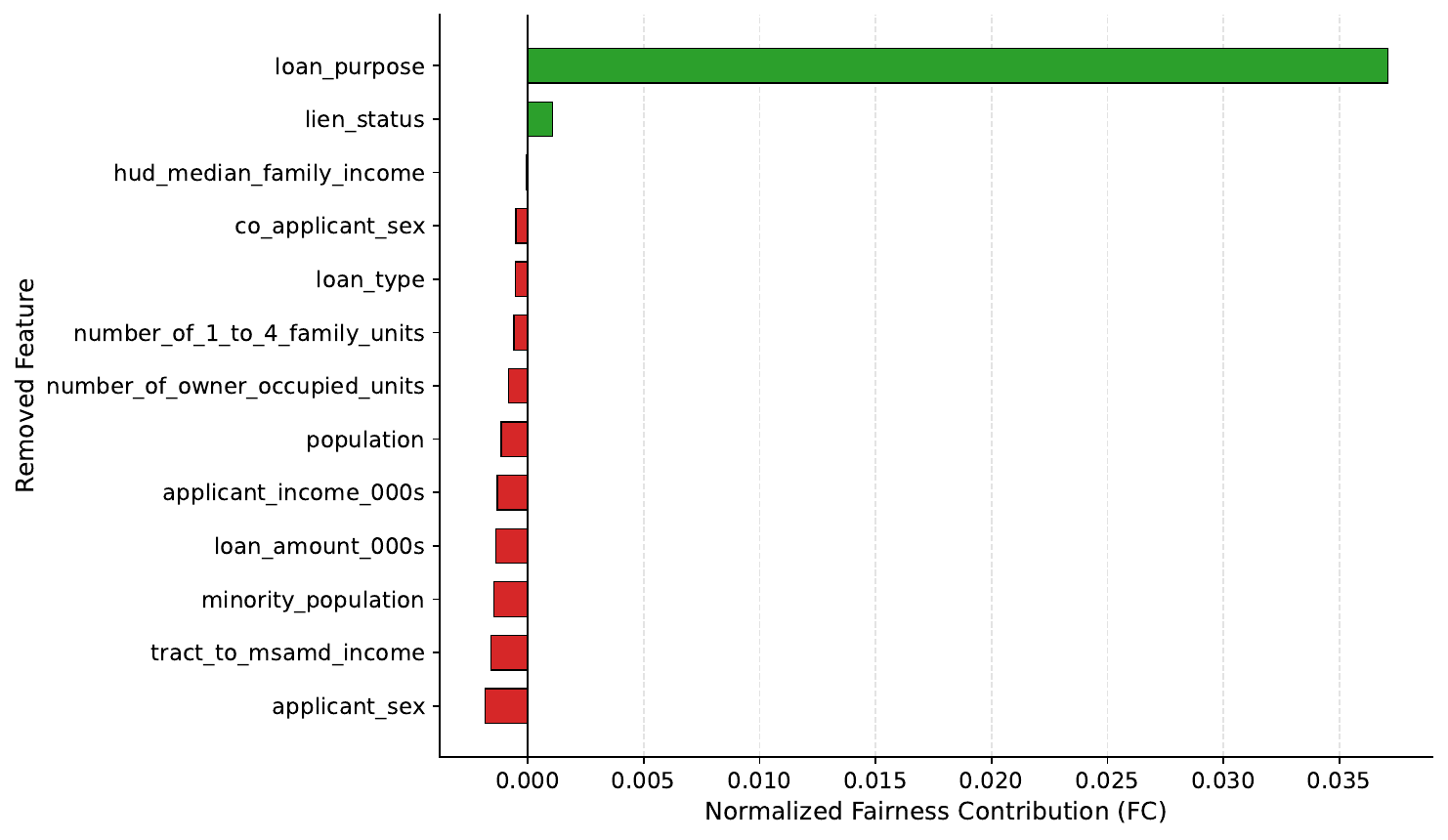}
    \caption{Gradient boosting}
    \label{fig:explainability_gb}
\end{subfigure}
\hfill
\begin{subfigure}[t]{0.48\textwidth}
    \centering
    \includegraphics[width=\linewidth]{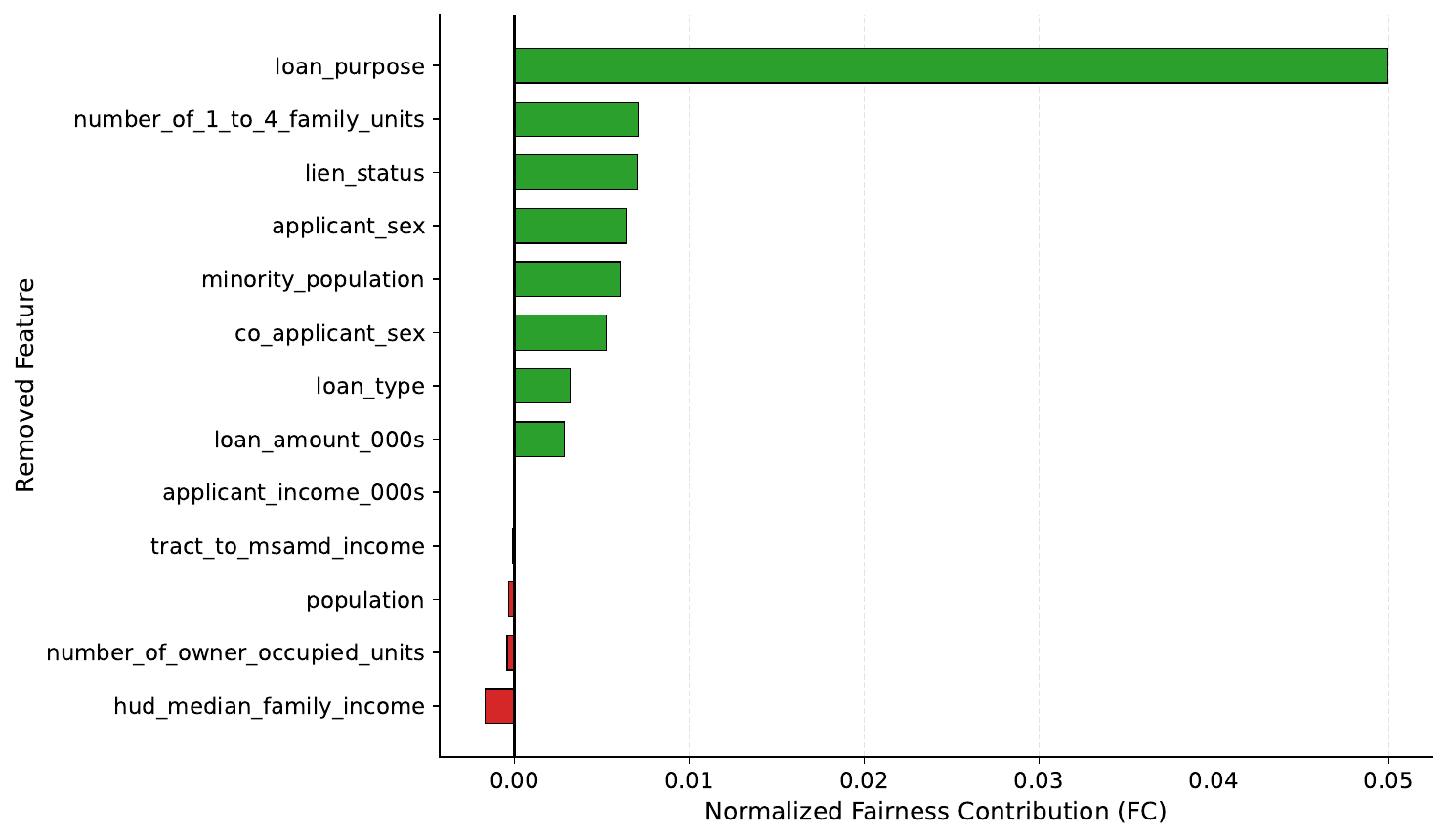}
    \caption{MLP}
    \label{fig:explainability_mlp}
\end{subfigure}
\caption{
Fairness explainability across features under imbalanced protected-group representation. The figures report the normalized fairness contribution
(\(FC_r\)) obtained by removing each predictor and re-estimating the
corresponding classifier. Positive contributions are shown in green and indicate that removing the corresponding feature increases RGF and therefore improves fairness. Negative contributions are shown in red and indicate that removing the feature decreases RGF and therefore worsens fairness.}
\label{fig:imbalanced_fairness_explainability}

\end{figure*}

The results of the feature removal analysis are reported in
Table~\ref{tab:imbalanced_explainability}, with the corresponding normalized fairness contributions shown in Figure~\ref{fig:imbalanced_fairness_explainability}. For each reduced model, the permutation test indicates whether the disparity across protected groups remains statistically significant after removing the feature.

Both Table~\ref{tab:imbalanced_explainability} and
Figure~\ref{fig:imbalanced_fairness_explainability} indicate that Loan purpose is the control variable that mostly affects fairness, consistently for all four models: its removal considerably improves it. However, unfairness remains significant at the $5\%$ level. The other variables have much smaller impacts, and their relevance depends on the considered model.

As shown above, improving fairness may come at the cost of predictive performance. For this reason, it is important to consider the fairness--performance trade-off induced by feature removal. Table~\ref{tab:imbalanced_explainability_performance}
 reports the changes in RGF and the normalized fairness contribution, along with the changes in ROC--AUC and PR--AUC, corresponding to each feature removal.

\begin{table*}[ht!]
\centering
\caption{Feature-removal fairness explainability results under imbalanced
protected-group representation}
\label{tab:imbalanced_explainability}

\begin{threeparttable}
\small
\setlength{\tabcolsep}{6pt}
\renewcommand{\arraystretch}{1.15}
\resizebox{0.97\textwidth}{!}{%
\begin{tabular}{
    l
    l
    c
    c
    c
    c
    c
    c
}
\toprule
Model
& Removed feature
& \(RGF_{\mathrm{full}}\)
& \(RGF_{-X_r}\)
& \(\Delta RGF_r\)
& \(FC_r\)
& CvM--\(T_{-X_r}\)
& \(p_{-X_r}\) \\
\midrule

Logistic regression
& loan\_purpose
& 0.6630 & 0.6898 & 0.0268 & 0.0796 & 137.3899 & \textbf{0.0005} \\

& minority\_population
& 0.6630 & 0.6601 & -0.0029 & -0.0044 & 64.4720 & \textbf{0.0005} \\

& co\_applicant\_sex
& 0.6630 & 0.6642 & 0.0012 & 0.0036 & 79.9383 & \textbf{0.0005} \\

& loan\_type
& 0.6630 & 0.6618 & -0.0012 & -0.0018 & 75.1155 & \textbf{0.0005} \\

& number\_of\_owner\_occupied\_units
& 0.6630 & 0.6635 & 0.0005 & 0.0015 & 80.2181 & \textbf{0.0005} \\

\midrule

Random forest
&loan\_purpose
& 0.6705 & 0.7538 & 0.0834 & 0.2530 & 40.8801 & \textbf{0.0005} \\

&minority\_population
& 0.6705 & 0.6552 & -0.0153 & -0.0227 & 12.5181 & \textbf{0.0005} \\

& hud\_median\_family\_income
& 0.6705 & 0.6736 & 0.0032 & 0.0096 & 19.2568 & \textbf{0.0005} \\

& loan\_amount\_000s
& 0.6705 & 0.6663 & -0.0041 & -0.0062 & 16.4233 & \textbf{0.0005} \\

& number\_of\_owner\_occupied\_units
& 0.6705 & 0.6713 & 0.0009 & 0.0026 & 19.7936 & \textbf{0.0005} \\

\midrule

Gradient boosting
& loan\_purpose
& 0.6573 & 0.6700 & 0.0127 & 0.0371 & 93.5549 & \textbf{0.0005} \\

& applicant\_sex
& 0.6573 & 0.6561 & -0.0012 & -0.0018 & 69.6440 & \textbf{0.0005} \\

& tract\_to\_msamd\_income
& 0.6573 & 0.6563 & -0.0010 & -0.0016 & 69.7053 & \textbf{0.0005} \\

& minority\_population
& 0.6573 & 0.6563 & -0.0010 & -0.0015 & 67.6854 & \textbf{0.0005} \\

& loan\_amount\_000s
& 0.6573 & 0.6564 & -0.0009 & -0.0014 & 67.6046 & \textbf{0.0005} \\

\midrule

MLP
& loan\_purpose
& 0.6559 & 0.6731 & 0.0172 & 0.0499 & 89.9525 & \textbf{0.0005} \\

& number\_of\_1\_to\_4\_family\_units
& 0.6559 & 0.6584 & 0.0024 & 0.0071 & 72.4208 & \textbf{0.0005} \\

& lien\_status
& 0.6559 & 0.6584 & 0.0024 & 0.0070 & 72.7879 & \textbf{0.0005} \\

& applicant\_sex
& 0.6559 & 0.6581 & 0.0022 & 0.0064 & 65.6799 & \textbf{0.0005} \\

& minority\_population
& 0.6559 & 0.6580 & 0.0021 & 0.0061 & 71.7725 & \textbf{0.0005} \\

\bottomrule
\end{tabular}
}

\vspace{2mm}

\begin{minipage}{0.94\textwidth}
\scriptsize
\textit{Notes:}
For each classifier, the Table reports the five features with the largest
absolute normalized fairness contribution, \(|FC_r|\).
\(RGF_{\mathrm{full}}\) denotes the RGF of the full model and
\(RGF_{-X_r}\) the RGF obtained after removing feature \(r\) and re-estimating the model. The change in fairness is
\(\Delta RGF_r=RGF_{-X_r}-RGF_{\mathrm{full}}\); hence, positive values indicate that removal of feature \(r\) improves fairness, whereas negative values indicate that removal worsens fairness.
\(FC_r\) denotes the normalized fairness contribution of feature \(r\).
CvM--\(T_{-X_r}\) and \(p_{-X_r}\) refer to the centered Cramér--von Mises statistic and protected-group permutation test for the reduced model, respectively. Bold \(p\)-values denote statistical significance at the 5\% level.
\end{minipage}

\end{threeparttable}
\end{table*}

\begin{table*}[ht!]
\centering
\caption{Main fairness contributions of individual features and associated changes in predictive performance under imbalanced protected-group representation}
\label{tab:imbalanced_explainability_performance}

\begin{threeparttable}
\small
\renewcommand{\arraystretch}{1.10}

\resizebox{0.97\textwidth}{!}{%
\begin{tabular}{
    l
    l
    c
    c
    c
    c
}
\toprule
Model
& Removed feature
& \(\Delta RGF_r\)
& \(FC_r\)
& \(\Delta\) ROC--AUC
& \(\Delta\) PR--AUC \\
\midrule

Logistic regression
& loan\_purpose
& \(0.0268\,\uparrow\)
& 0.0796
& \(-0.0757\,\downarrow\)
& \(-0.0560\,\downarrow\) \\

& minority\_population
& \(-0.0029\,\downarrow\)
& -0.0044
& \(-0.0028\,\downarrow\)
& \(-0.0082\,\downarrow\) \\

& co\_applicant\_sex
& \(0.0012\,\uparrow\)
& 0.0036
& \(-0.0033\,\downarrow\)
& \(-0.0058\,\downarrow\) \\

& loan\_type
& \(-0.0012\,\downarrow\)
& -0.0018
& \(-0.0021\,\downarrow\)
& \(0.0010\,\uparrow\) \\

&number\_of\_owner\_occupied\_units
& \(0.0005\,\uparrow\)
& 0.0015
& \(-0.0003\,\downarrow\)
& \(0.0004\,\uparrow\) \\

\midrule

Random forest
& loan\_purpose
& \(0.0834\,\uparrow\)
& 0.2530
& \(-0.0538\,\downarrow\)
& \(-0.0458\,\downarrow\) \\

& minority\_population
& \(-0.0153\,\downarrow\)
& -0.0227
& \(-0.0011\,\downarrow\)
& \(-0.0022\,\downarrow\) \\

&hud\_median\_family\_income
& \(0.0032\,\uparrow\)
& 0.0096
& \(-0.0048\,\downarrow\)
& \(-0.0056\,\downarrow\) \\

& loan\_amount\_000s
& \(-0.0041\,\downarrow\)
& -0.0062
& \(-0.0159\,\downarrow\)
& \(-0.0318\,\downarrow\) \\

& number\_of\_owner\_occupied\_units
& \(0.0009\,\uparrow\)
& 0.0026
& \(0.0002\,\uparrow\)
& \(0.0000\,=\) \\

\midrule

Gradient boosting
& loan\_purpose
& \(0.0127\,\uparrow\)
& 0.0371
& \(-0.0513\,\downarrow\)
& \(-0.0410\,\downarrow\) \\

& applicant\_sex
& \(-0.0012\,\downarrow\)
& -0.0018
& \(-0.0001\,\downarrow\)
& \(0.0001\,\uparrow\) \\

& tract\_to\_msamd\_income
& \(-0.0010\,\downarrow\)
& -0.0016
& \(0.0017\,\uparrow\)
& \(0.0028\,\uparrow\) \\

& minority\_population
& \(-0.0010\,\downarrow\)
& -0.0015
& \(0.0010\,\uparrow\)
& \(0.0012\,\uparrow\) \\

& loan\_amount\_000s
& \(-0.0009\,\downarrow\)
& -0.0014
& \(-0.0162\,\downarrow\)
& \(-0.0348\,\downarrow\) \\

\midrule

MLP
& loan\_purpose
& \(0.0172\,\uparrow\)
& 0.0499
& \(-0.0555\,\downarrow\)
& \(-0.0472\,\downarrow\) \\

& number\_of\_1\_to\_4\_family\_units
& \(0.0024\,\uparrow\)
& 0.0071
& \(-0.0004\,\downarrow\)
& \(0.0004\,\uparrow\) \\

& lien\_status
& \(0.0024\,\uparrow\)
& 0.0070
& \(-0.0018\,\downarrow\)
& \(-0.0030\,\downarrow\) \\

& applicant\_sex
& \(0.0022\,\uparrow\)
& 0.0064
& \(-0.0002\,\downarrow\)
& \(0.0007\,\uparrow\) \\

& minority\_population
& \(0.0021\,\uparrow\)
& 0.0061
& \(-0.0021\,\downarrow\)
& \(-0.0033\,\downarrow\) \\

\bottomrule
\end{tabular}%
}
\vspace{2mm}

\begin{minipage}{0.97\textwidth}
\scriptsize
\textit{Notes:}
For each classifier, the five predictors with the largest absolute normalized fairness contributions, \(|FC_r|\), are reported.
\(\Delta RGF_r=RGF_{-X_r}-RGF_{\mathrm{full}}\) measures the change in
fairness after removing predictor \(r\) and re-estimating the classifier.
For \(\Delta RGF_r\), \(\uparrow\) indicates improved fairness,
\(\downarrow\) indicates worsened fairness, and \(=\) indicates no change at the reported precision. \(FC_r\) denotes the normalized fairness contribution. \(\Delta\)ROC--AUC and \(\Delta\)PR--AUC measure changes in predictive performance relative to the full model; \(\uparrow\) denotes an improvement, \(\downarrow\) a deterioration, and \(=\) no change at the reported precision.
\end{minipage}

\end{threeparttable}
\end{table*}

Table~\ref{tab:imbalanced_explainability_performance} could 
distinguish features whose removal yields a fairness gain at limited
predictive cost, from those for which improved fairness is obtained at the
expense of a substantial loss in classification performance. The evidence in the Table shows that the variable whose removal yields the largest fairness gain, loan\_purpose, also has a high removal cost. Other variables, less important from a fairness gain, also have a negligible elimination cost: this is the case, for example, for the co\_applicant\_sex variable.

From a credit lending perspective, our analysis complements the findings of \cite{giudici2025safe}. Using the same HMDA data to assess accuracy, explainability, and robustness, the authors identify logistic regression as
the model with the strongest integrated performance. The fairness analysis adds a further dimension to this comparison. In particular, the results show that descriptive fairness measures should be considered alongside formal inference, especially under pronounced imbalance in protected group representation. Random Forest provides a clear example: it records the highest AURGF and the lowest centered Cramér--von Mises statistic among the four models, indicating the smallest distributional disparity. Nevertheless, the permutation test remains significant, so the model cannot be regarded as fair. This distinction illustrates the value of complementing descriptive fairness measures with statistical inference when comparing models for credit lending applications.


\subsection{HMDA data, balanced}
\label{subsec:balanced_group_fairness}

For robustness, we now balance the HMDA data and repeat the analysis.  To construct a group-balanced setting for the fairness assessment,
the majority group was randomly undersampled, without replacement, to match the size of the minority group, using a fixed seed. The resulting dataset is equally divided between the two protected groups. Following undersampling, the target distribution
comprised 75.83\% negative outcomes and 24.17\%
positive outcomes. Because sampling was conditioned
exclusively on protected-group membership, the change in outcome
prevalence arose from the association between group membership and the target in the original data; the outcome itself was not directly
resampled. The protected attribute was excluded from the predictor
matrix and retained solely for the subsequent fairness audit.

We tuned hyperparameters separately under the two sampling designs. We first tuned on the original imbalanced dataset for the full-sample analysis. After protected-group undersampling, we repeated the full tuning procedure on the balanced dataset. The configurations described below therefore apply exclusively to the balanced-group experiment and were not transferred from the full-sample models.

Logistic regression was selected with \(C=73.9227\), a convergence
tolerance of \(8.17\times10^{-5}\), and no class weighting. The random
forest comprised 700 trees and employed bootstrap sampling, the
log-loss splitting criterion, a maximum depth of 8, a maximum-feature
proportion of \(0.8\), a minimum leaf size of 3, and a minimum split
size of 13. Gradient boosting was specified with 314 estimators, a
learning rate of \(0.0327\), a maximum depth of 5, square-root feature
sampling, a subsampling rate of \(0.8323\), a minimum leaf size of 17,
and a minimum split size of 9. The MLP comprised three hidden layers
containing \(128\), \(64\), and \(32\) neurons, respectively, with
hyperbolic-tangent activation, a batch size of 512, an \(L_2\)
regularization parameter of \(2.46\times10^{-4}\), an initial learning
rate of \(2.15\times10^{-4}\), and a no-improvement stopping window of
10 iterations. We evaluated the final models on a held-out test partition. The resulting predictive performance and fairness
estimates are reported in Table~\ref{tab:balanced_model_comparison}.

\begin{table*}[ht!]
\centering
\caption{Predictive performance and rank-based fairness results after
protected-group undersampling}
\label{tab:balanced_model_comparison}

\begin{threeparttable}
\small
\setlength{\tabcolsep}{5pt}
\renewcommand{\arraystretch}{1.15}
\resizebox{0.97\textwidth}{!}{%
\begin{tabular}{
    l
    cc
    cc
    c
    c
    c
    c
    c
}
\toprule
& \multicolumn{2}{c}{Predictive performance}
& \multicolumn{4}{c}{Rank-based fairness}
& \multicolumn{2}{c}{Statistical inference}
& \\

\cmidrule(lr){2-3}
\cmidrule(lr){4-7}
\cmidrule(lr){8-9}

Model
& ROC--AUC
& PR--AUC
& RGD
& RGF
& \makecell{RGF\\95\% CI}
& AURGF
& CvM--$T$
& Permutation \(p\)-value
& \(q_{\min}\) \\
\midrule

Logistic regression
& 0.7028
& 0.4252
& 0.1906
& 0.8094
& [0.7930, 0.8234]
& 0.6023
& 70.6250
& \textbf{0.0005}
& 0.0101 \\

Random forest
& 0.7296
& 0.4817
& 0.2123
& 0.7877
& [0.7607, 0.8152]
& 0.5492
& \textbf{28.0435}
& \textbf{0.0005}
& 0.0101 \\

Gradient boosting
& \textbf{0.7327}
& \textbf{0.4903}
&\textbf{ 0.1744}
& \textbf{0.8256}
& [0.8077, 0.8409]
& \textbf{0.6790}
& 65.3432
& \textbf{0.0005}
& 0.0101 \\

MLP
& 0.7122
& 0.4541
& 0.1892
& 0.8108
& [0.7918, 0.8259]
& 0.6417
& 67.8869
& \textbf{0.0005}
& 0.0101 \\

\bottomrule
\end{tabular}
}
\vspace{2mm}

\begin{minipage}{0.94\textwidth}
\scriptsize
\textit{Notes:}
Results are computed after random undersampling of the majority protected group, yielding two equally sized groups. Values in brackets are 95\% group-stratified bootstrap confidence intervals for RGF.
CvM--\(T\) denotes the centered observed Cramér--von Mises statistic.
Permutation \(p\)-values are based on protected-label
permutations. Bold entries indicate the best predictive
performance,  the lowest RGD and CvM--\(T\), and the highest RGF and AURGF, whereas bold \(p\)-values denote statistical significance at the 5\% level. The quantity \(q_{\min}\) denotes the smallest top-burden fraction used to construct the partial RGF curve.
\end{minipage}

\end{threeparttable}
\end{table*}

Table~\ref{tab:balanced_model_comparison} shows a different model ranking after undersampling the majority protected group. On the original imbalanced sample, Random Forest records the highest RGF and AURGF, whereas Gradient Boosting achieves the best classification performance. After undersampling, Gradient Boosting ranks first in both predictive performance and descriptive fairness, with the highest ROC--AUC, PR--AUC, RGF, and AURGF.

This change is closely related to the composition of the outcome variable.
Approved loans are more strongly represented in the privileged group (majority), so undersampling this group reduces the share of approvals and increases the proportion of declined applications from 18.37\% to 24.17\%. The resulting target is therefore less imbalanced, which provides more favorable conditions for classification and, importantly, a more balanced setting for interpreting RGF and AURGF. This is consistent with the simulation results, which showed that severe protected-group imbalance can distort descriptive fairness measures. Gradient Boosting also produced the lowest prediction errors in the unfair simulation scenario, where the permutation test remained non-significant over the initial retained fractions. Because the proposed fairness measures are constructed from prediction error burdens, this result coherently explains the joint improvement in predictive performance and descriptive fairness observed after undersampling. Nevertheless, the permutation \(p\)-value remains significant in the balanced HMDA sample, showing that the statistical evidence of disparity persists despite the more favorable RGF and AURGF values.

\subsection{Comparison with classical fairness metrics}

Table~\ref{tab:imbalanced_classical_fairness} highlights the distinction
between the conventional fairness criteria and the perspective developed in this paper. Logistic regression records the smallest SPD, EOD, FPR difference, and PPD, although it also has the lowest PR--AUC (\(0.3198\)). Random Forest, by contrast, records the lowest DI (\(1.6852\)), while performing less favorably on the other classical criteria. These conflicting signals align with the documented sensitivity of classical fairness measures to class and protected-group imbalance \citep{brzezinski2024properties}. They do not allow a conclusive fairness evaluation for the HMDA data.

The proposed framework addresses fairness at a different level. Rather than assessing parity from classification outcomes alone, RGF and AURGF evaluate how error burdens are distributed across protected groups, thereby linking fairness to predictive reliability. This perspective explains why Random Forest, despite its less favorable classical metrics, attains the highest RGF and AURGF. However, the simulated data application shows that severe group imbalance can affect descriptive rank-based measures. Our proposed permutation test provides the necessary correction, leading to the lowest centered Cramér--von Mises statistic for the Random Forest model. The test rejects the fairness null hypotheses for all four classifiers, providing conclusive evidence that unfairness in the HMDA application is native to the data and does not depend on the machine learning model considered.
We also note that classical metrics do not identify which explanatory variables, if any, explain unfairness. Our framework instead, as shown in \ref{explain}, identifies loan\_purpose as the most important feature.

Taken together, the results from comparing our proposal with classical fairness metrics show that trustworthy AI assessment requires moving beyond parity measures toward a joint evaluation of predictive accuracy, fairness, and statistical evidence at the individual level.

\begin{table}[ht!]
\centering
\caption{Classical fairness metrics on the original dataset with imbalanced protected groups}
\label{tab:imbalanced_classical_fairness}

\begin{threeparttable}
\footnotesize
\renewcommand{\arraystretch}{1.15}

\resizebox{\textwidth}{!}{%
\begin{tabular}{
    l
    c
    c
    c
    c
    c
}
\toprule
Model
& SPD
& DI
& EOD
& FPR difference
& PPD \\
\midrule

Logistic regression
& 0.0150
& 11.4081
& 0.0203
& 0.0118
& 0.0093 \\

Random forest
& 0.2071
& 1.6852
& 0.1566
& 0.1627
& 0.1284 \\

Gradient boosting
& 0.0706
& 2.7861
& 0.0766
& 0.0419
& 0.0149 \\

MLP
& 0.0388
& 2.4498
& 0.0391
& 0.0195
& 0.0298 \\

\bottomrule
\end{tabular}%
}
\vspace{2mm}

\begin{minipage}{0.94\textwidth}
\scriptsize
\textit{Notes:}
SPD denotes statistical parity difference, DI disparate impact, EOD equal
opportunity difference, FPR difference, the difference in false positive rates, and PPD predictive parity difference.
\end{minipage}

\end{threeparttable}
\end{table}


\section{Conclusions}
\label{sec:conclusions}

This paper contributes to the literature on fairness in algorithmic decisions, specifically in credit lending, by proposing a framework that quantifies, localizes, and explains disparities in machine learning models.

It does so by means of four related methodologies:
i) the definition of a SAFE-AI Rank Graduation Fairness metric (RGF), able to quantify disparities in prediction error burdens; ii) the development of a permutation test, based on a centered Cram\'er--von Mises (CvM) statistic, to assess their statistical significance; iii) the construction of the Area Under the Rank Graduation Fairness Curve (AURGF), to determine where disparities are concentrated along the prediction error distribution; and iv) a fairness explainability procedure based on feature removal, to identify the variables associated with improvements or deteriorations in fairness.

The simulation study establishes an important distinction between descriptive and inferential fairness assessment. Under severe protected group imbalance, RGF and AURGF can be affected by group composition and may even reverse the descriptive fairness ranking. The proposed centered CvM permutation test addresses this issue by correctly distinguishing fair and unfair generating mechanisms. The simulation therefore shows that RGF and AURGF should be accompanied by statistical inference when protected groups are substantially imbalanced.

The application to the HMDA data provides further evidence. Unlike classical parity criteria, which produce model rankings at the group level, the proposed framework evaluates fairness through the distribution of prediction errors. This establishes a direct link between
fairness and predictive accuracy, an important consideration in algorithmic decisions and credit scoring, where it is important to identify for which individuals unfairness arises, and for which reasons.  The
ensemble tree models perform particularly well in this respect. Random Forest provides the strongest descriptive fairness results on the original
imbalanced sample, whereas, after protected group undersampling, Gradient Boosting achieves both the highest predictive performance and the highest RGF and AURGF. 

The proposed statistical tests better qualify these descriptive rankings.
The fairness null is rejected for all four classifiers, irrespective of
whether the model is a logistic regression,  an ensemble tree, or a neural network. The persistence of the result across learning architectures suggests the observed disparity is not specific to any one classifier. Rather, it is consistent with systematic group differences embedded in the lending data. This interpretation aligns with \cite{bhutta2025much}, who show that racial differences in mortgage decisions can persist in race-blind automated decision systems, regardless of the model used.  

The fairness explainability analysis reinforces this conclusion. Loan purpose is consistently the most influential variable, and removing it yields the largest fairness improvements, though at a non-negligible cost in predictive performance. In addition, disparity stays statistically significant after its exclusion. Unfairness in the HMDA application therefore cannot be removed by eliminating predictors. The findings illustrate the value of combining fairness explainability with inference rather than interpreting improvements in descriptive fairness as evidence that disparity has been eliminated.

We finally remark that the proposed methodology is quite general. By defining fairness in terms of the distribution of prediction errors, it establishes a direct connection between fairness, predictive accuracy, and explainability. Although in this paper we have considered the credit lending data, the most analyzed algorithmic decision setting, likely because of its highly regulated environment, the methodology can be equally applied to similar algorithmic decisions: for example, in student admission, job candidate ranking, medical diagnosis, cyber attack detection, and judicial decisions. Future work could extend the application to these domains.

\section*{Data and reproducibility}

The empirical analysis uses 2017 Home Mortgage Disclosure Act (HMDA)
data for New York State, obtained from the Consumer Financial Protection
Bureau (CFPB). The data are publicly available through the
\href{https://www.consumerfinance.gov/data-research/hmda/historic-data/?geo=ny&records=all-records&field_descriptions=labels}
{CFPB HMDA historic data repository}.

The Python code employed for the empirical analysis in this paper is
available from the authors upon request.

\bibliographystyle{elsarticle-harv}
\bibliography{cas-refs}






\appendix
\section{Theoretical Foundation of the Permutation \(p\)-Value}
\label{app:pvalue_theory}

\subsection{Null Hypothesis and Test Statistic}

Let
\begin{equation}
\mathbf{z}=(z_1,\ldots,z_n)
\end{equation}
denote the observed error burdens, and let
\begin{equation}
\mathbf{g}=(g_1,\ldots,g_n)
\end{equation}
denote the corresponding protected-group labels. The permutation test considers the null hypothesis:
\begin{equation}
H_0:
\mathbf{g}
\text{ is exchangeable conditional on }
\mathbf{z}
\end{equation}

Under \(H_0\), protected-group membership carries no systematic information about the distribution of error burdens. The error values and their ordering are therefore held fixed, while the group labels are reassigned subject to the observed group sizes. For groups \(g\) and \(h\), the observed Cramér--von Mises statistic is:
\begin{equation}
T_{gh}^{obs}
=
\sum_{k=1}^{n}
\left[
\widetilde{E}_{g}(t_k)
-
\widetilde{E}_{h}(t_k)
\right]^2
\end{equation}

Larger values indicate greater separation between the completed group curves and hence provide stronger evidence against \(H_0\). For notational simplicity, the pairwise subscripts are omitted below.

\subsection{Exact Randomization Distribution}

Let \(\mathcal{P}\) denote the set of all admissible assignments of the group labels that preserve the observed group sizes. For each assignment \(\tau\in\mathcal{P}\), the completed group curves are recomputed, and the corresponding statistic is obtained as:
\begin{equation}
T^{\tau}
=
\sum_{k=1}^{n}
\left[
\widetilde{E}_{g}^{\tau}(t_k)
-
\widetilde{E}_{h}^{\tau}(t_k)
\right]^2
\end{equation}

When all admissible assignments can be enumerated, the exact permutation \(p\)-value is:
\begin{equation}
p_{\mathrm{exact}}
=
\frac{
\displaystyle
\sum_{\tau\in\mathcal{P}}
I\left(
T^{\tau}\geq T^{obs}
\right)
}
{|\mathcal{P}|}
\end{equation}

For two groups of sizes \(n_g\) and \(n_h\), the number of admissible assignments is:
\begin{equation}
|\mathcal{P}|
=
\binom{n_g+n_h}{n_g}
\end{equation}

Thus, \(p_{\mathrm{exact}}\) is the proportion of admissible label assignments producing a curve separation at least as large as the observed separation.

\subsection{Monte Carlo Permutation \(p\)-Value}

Exhaustive enumeration may be computationally infeasible when the number of admissible assignments is large. In that case, \(B\) admissible permutations are sampled, and
\begin{equation}
M
=
\sum_{b=1}^{B}
I\left(
T^{(b)}
\geq
T^{{obs}}
\right)
\end{equation}
denotes the observed number of permutation statistics at least as large as \(T^{{obs}}\). The corrected Monte Carlo permutation \(p\)-value is:
\begin{equation}
p_{\mathrm{perm}}
=
\frac{M+1}{B+1}
\label{eq:corrected_perm_pvalue}
\end{equation}

The correction is based on the number of exceedances expected under \(H_0\) when only \(B\) permutations are performed. Let \(M^{\ast}\) denote the random number of exceedances among \(B\) Monte Carlo permutations, with \(M\) its observed realization, and let \(p_{\infty}\) denote the underlying probability that a statistic generated under \(H_0\) is at least as large as the observed statistic. Conditional on \(p_{\infty}\):
\begin{equation}
M^{\ast}\mid p_{\infty}
\sim
\operatorname{Binomial}(B,p_{\infty})
\end{equation}

Under \(H_0\), the observed statistic can occupy any position in the continuous null distribution. Therefore, \(p_{\infty}\) is uniformly distributed over \([0,1]\). Consequently, for \(m=0,\ldots,B\):
\begin{align}
\Pr_{H_0}(M^{\ast}=m)
&=
\int_0^1
\binom{B}{m}
p_{\infty}^{m}
(1-p_{\infty})^{B-m}
\,dp_{\infty}
\nonumber\\
&=
\binom{B}{m}
\frac{m!(B-m)!}{(B+1)!}
=
\frac{1}{B+1}
\end{align}

The integral equals \(1/(B+1)\) for every possible value of \(m\). Therefore, under \(H_0\), each possible exceedance count from \(0\) to \(B\) has the same probability. After observing \(M\) exceedances, the probability of obtaining \(M\) or fewer exceedances under \(H_0\) is:
\begin{equation}
\Pr_{H_0}(M^{\ast}\leq M)
=
\sum_{m=0}^{M}
\frac{1}{B+1}
=
\frac{M+1}{B+1}
\end{equation}

The denominator \(B+1\) reflects the \(B+1\) possible exceedance counts,
\(0,\ldots,B\), whereas the numerator \(M+1\) reflects the \(M+1\) outcomes contained in the event \(M^{\ast}\leq M\), namely \(0,\ldots,M\). When ties occur, the non-strict comparison $T^{(b)}
\geq T^{\mathrm{obs}}$ counts permutation statistics equal to the observed statistic as equally extreme. This convention may increase the resulting \(p\)-value and therefore yields a conservative test. Accordingly:
\begin{equation}
\Pr_{H_0}
\left(
p_{\mathrm{perm}}\leq\alpha
\right)
\leq
\alpha
\end{equation}

This establishes finite-sample control of the type I error rate. Because \(M\in\{0,\ldots,B\}\), the attainable \(p\)-values are:
\begin{equation}
\frac{1}{B+1},
\frac{2}{B+1},
\ldots,
1,
\end{equation}
and the minimum attainable value is:
\begin{equation}
p_{\min}
=
\frac{1}{B+1}
\end{equation}

Consequently, a finite Monte Carlo permutation procedure cannot produce a zero \(p\)-value. If none of the sampled permutation statistics is at least as large as \(T^{obs}\), then \(M=0\) and \(p_{\mathrm{perm}} = 1/(B+1)\).






\end{document}